\documentclass[conference]{IEEEtran}
\IEEEoverridecommandlockouts
\usepackage{cite}
\usepackage{amsmath,amssymb,amsfonts}
\usepackage{graphicx}
\usepackage{hyperref}
\usepackage{textcomp}
\usepackage[dvipsnames]{xcolor}
\usepackage{bm}
\usepackage{amsthm}
\usepackage{subfigure}
\usepackage[linesnumbered,ruled,noend]{algorithm2e} 
\usepackage{algorithmic}
\usepackage{multirow}
\usepackage{float}
\usepackage{siunitx}
\usepackage{wrapfig}
\usepackage{dblfnote}
\usepackage{cancel}
\def\BibTeX{{\rm B\kern-.05em{\sc i\kern-.025em b}\kern-.08em
    T\kern-.1667em\lower.7ex\hbox{E}\kern-.125emX}}
\begin{document}

\title{Climate Modeling with Neural Diffusion Equations}

\author{\IEEEauthorblockN{Jeehyun Hwang$^1$, Jeongwhan Choi$^1$, Hwangyong Choi$^1$, Kookjin Lee$^2$, Dongeun Lee$^3$, Noseong Park$^1$}
\textit{Yonsei University$^1$, Seoul, South Korea}\\
\textit{Arizona State University$^2$, Tempe, AZ, USA}\\
\textit{Texas A\&M University--Commerce$^3$, Commerce, TX, USA}
}


\maketitle

\begin{abstract}
Owing to the remarkable development of deep learning technology, there have been a series of efforts to build deep learning-based climate models. Whereas most of them utilize recurrent neural networks and/or graph neural networks, we design a novel climate model based on the two concepts, the neural ordinary differential equation (NODE) and the diffusion equation. Many physical processes involving a Brownian motion of particles can be described by the diffusion equation and as a result, it is widely used for modeling climate. On the other hand, neural ordinary differential equations (NODEs) are to learn a latent governing equation of ODE from data. In our presented method, we combine them into a single framework and propose a concept, called \emph{neural diffusion equation} (NDE). Our NDE, equipped with the diffusion equation and one more additional neural network to model inherent uncertainty, can learn an appropriate latent governing equation that best describes a given climate dataset. In our experiments with two real-world and one synthetic datasets and eleven baselines, our method consistently outperforms existing baselines by non-trivial margins.
\end{abstract}


\section{Introduction}
Deep learning-based climate modeling (or weather forecasting) is an emerging topic and a brand-new application area~\cite{zaytar2016sequence, shi2015convolutional,shi2017deep, liu2016application, racah2016extremeweather, kurth2018exascale, cheng2018ensemble, cheng2018neural, hossain2015forecasting, ren2021deep, tekin2021spatio}. Owing to the recent advancement of the differential equation-inspired deep learning~\cite{NIPS2018_7892, greydanus2019hamiltonian, finzi2020simplifying, cranmer2020lagrangian, lutter2019deep}, this specific topic is gathering much attention from the research community.

The seminal paper, titled neural ordinary differential equations (NODEs), discovered that residual networks are equivalent to the explicit Euler method to solve ODE problems~\cite{NIPS2018_7892}. Therefore, training residual networks is solving ODE problems specialized in image classification, according to them.

In NODEs, more specifically, we solve an integral problem of $\bm{h}(t_1) = \bm{h}(t_0) + \int_{t_0}^{t_1}f(\bm{h}(t),t;\bm{\theta}_f) dt$, where $\bm{h}$ is a vector that contains a set of values that change over time $t \in [0,T]$ and $f(\bm{h}(t),t;\bm{\theta}_f) = \frac{d\bm{h}(t)}{dt}$. In other words, the multi-dimensional vector $\bm{h}(t_1)$ at time $t_1 > t_0$ is calculated by adding the sum of the changes in $[t_0, t_1]$ to $\bm{h}(t_0)$. The key of NODEs is the neural network $f$ parameterized by $\bm{\theta}_f$, which is trained from data.

In the case of climate modeling, an area to model is frequently abstracted by a grid network (or a small-world network) and each grid cell corresponds to an element in $\bm{h}(t)$~\cite{seo2019differentiable}. We then learn $\bm{\theta}_f$ from data to minimize a climate modeling loss function. We typically use the mean squared error (MSE) as a loss function since climate modeling or weather forecasting is a regression problem.

Whereas various recurrent neural network models and regression models can be used for this purpose, we propose to use the diffusion equation~\cite{strauss2007partial} under the regime of NODEs to model the physical dynamics governing the spread of temperature, air, etc. As a matter of fact, the diffusion equation is one of the most popular differential equations for climate modeling~\cite{stocker2011introduction,larwa2019heat}.

\begin{figure}
    \centering
    \subfigure[]{\includegraphics[width=\columnwidth]{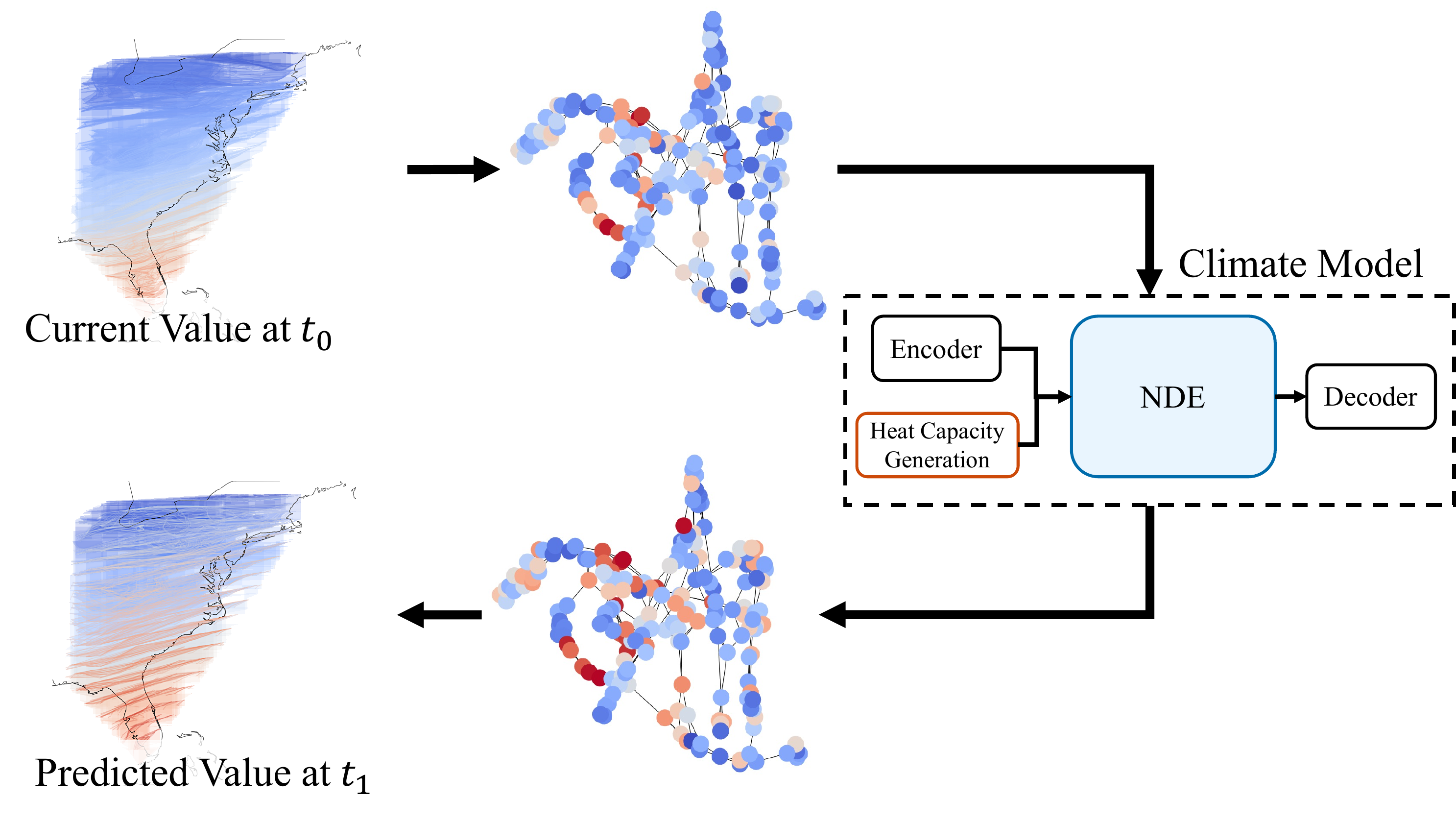}}
    \subfigure[]{\includegraphics[width=\columnwidth]{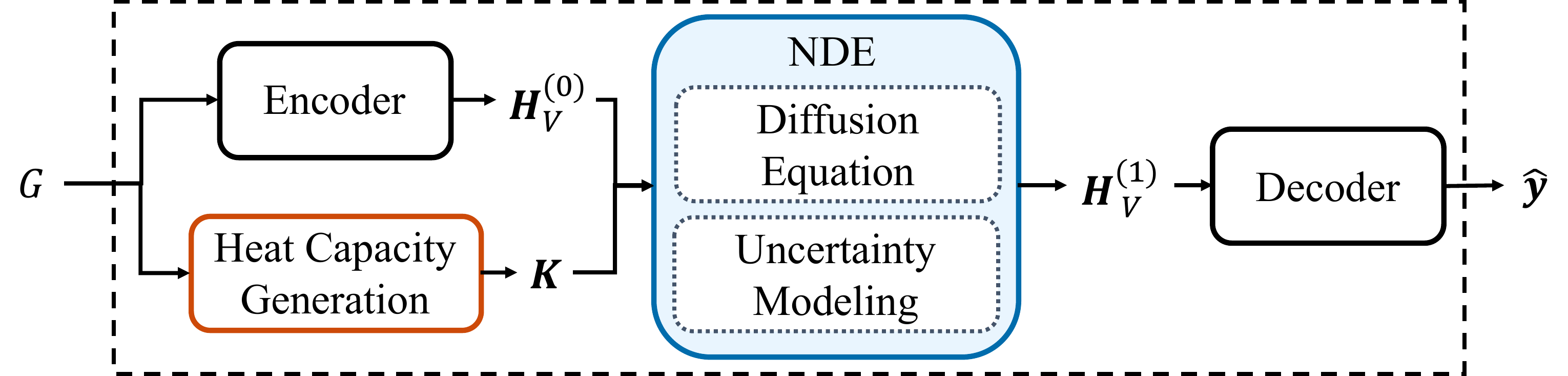}}
    \caption{The overall workflow of our proposed \emph{neural diffusion equation} (NDE). NDE stands for our proposed method as well as its key layer. (a) The weather stations (and their sensing values) are represented as a graph and our NDE predicts their future values. (b) Given a graph $G=(V,E)$ annotated with node and edge features, the core NDE layer evolves $\bm{H}_V^{(0)}$ to $\bm{H}_V^{(1)}$, followed by a decoding layer which produces predictions. The NDE layer consists of two parts: i) the diffusion equation and ii) the neural network-based uncertainty model $f$.}
    \label{fig:archi}
\end{figure}

The diffusion equation over a grid network uses the discrete Laplacian operator which is written as $\frac{d\bm{h}(t)}{dt} = -\Delta \bm{h}(t) = - k \bm{L}\bm{h}(t)$, where $k$ is a heat capacity coefficient, $\Delta$ means the discrete Laplacian operator, and $\bm{L}$ is the symmetrically normalized graph Laplacian matrix. The physical process in this equation is governed by the current values $\bm{h}(t)$ at time $t$ and the grid connectivity.

One thing worth mentioning in the above diffusion equation is that it does not have any parameters to learn since $\bm{L}$ and $\bm{h}(0)$, i.e., the initial climate conditions, are given by data. In this regard, it is a pure ODE rather than NODE. To further increase the accuracy, in our case we extend the definition of the time-derivative of $\bm{h}(t)$ to $\frac{d\bm{h}(t)}{dt} = - k \bm{L}\bm{h}(t) + f(\bm{h}(t),t;\bm{\theta}_f)$ --- we note that our model can be considered as a NODE since the time-derivative of $\bm{h}(t)$ is partially approximated by the neural network $f$. The neural network $f$ is to learn the uncertainty of real-world diffusion processes. Therefore, we solve $\bm{h}(t_1) = \bm{h}(t_0) + \int_{t_0}^{t_1}- k \bm{L}\bm{h}(t) + f(\bm{h}(t),t;\bm{\theta}_f) dt$, which we call \emph{neural diffusion equation} (NDE). We also consider the case that $k$ is not a scalar value but a matrix (see Section~\ref{sec:hc}). For simplicity but without loss of generality, we assume $k$ is a scalar in this section.

The overall workflow of our method is shown in Fig.~\ref{fig:archi}. In most cases, weather stations and their sensing values are converted into a graph annotate with node and/or edge features, and we use our neural diffusion equation (NDE) method to predict future climate-related factors. In our proposed method, the heat capacity coefficients are also trained from data and so is the neural network-based uncertainty model. Since real-world environments accompany uncertainties and noises incurred by human activities and so forth, our uncertainty modeling is one of the most important part in the method. The heat capacity defines how quickly (easily) two neighboring nodes interact with each other, which is not explicitly given to us in many cases. Therefore, we propose to learn from data. Another key component in our method is the heat equation. We also implement the exact heat equation since non-trivial parts in climate modeling can still be modeled by it.

In our experiments, we show that our method, featured by i) the heat equation, ii) the uncertainty modeling, and iii) the heat capacity generation, shows the best accuracy in all datasets. We use five datasets: two of them are synthetic and the other three are real-world datasets. The synthetic datasets include a grid and small-world network (with and without noises injected into the data) --- we note that climate models typically assume either a grid or a small-world network\footnote{Each research domain has its own preference on graph models. For instance, social networks typically assume scale-free networks.}. The real-world datasets includes the data collected in Los Angeles, San Diego, and the east side of the USA. We compare our method with eleven baseline methods and our proposed method NDE shows the best forecasting performance only except the multi-step forecasting in San Diego. Our contributions can be summarized as below:
\begin{enumerate}
    \item We design a climate forecasting model with the diffusion equation, the neural network-based uncertainty modeling, and the heat capacity generation methods.
    \item We conduct comprehensive experiments with synthetic and real-world datasets and compare with eleven baselines. Our proposed method, called \emph{neural diffusion equation} (NDE), shows the best forecasting performance only except one experimental case. Our method's forecasting errors are up to 51\% smaller than that of the best baseline method.
\end{enumerate}

\section{Related Work}
We introduce several base concepts in this section: i) neural ordinary differential equations, ii) diffusion equations, and iii) climate modeling.

\subsection{Neural Ordinary Differential Equations (NODEs)}\label{sec:node}
NODEs solve the following initial value problem (IVP), which involves an integral problem, to calculate $\bm{z}(t_1)$ from $\bm{z}(t_0)$~\cite{NIPS2018_7892}:
\begin{align}
    \bm{z}(t_1) = \bm{z}(t_0) + \int_{t_0}^{t_1}f(\bm{z}(t);\bm{\theta}_f)dt,
\end{align}where $f(\bm{z}(t);\bm{\theta}_f)$, which we call \emph{ODE function}, is a neural network to approximate $\dot{\bm{z}} \stackrel{\text{def}}{=} \frac{d \bm{z}(t)}{d t}$. To solve the integral problem, NODEs rely on ODE solvers, such as the explicit Euler method, the Dormand--Prince (DOPRI) method, and so forth~\cite{DORMAND198019}.

In general, ODE solvers discretize time variable $t$ and convert an integral into many steps of additions. For instance, the explicit Euler method can be written as follows in a step:
\begin{align}\label{eq:euler}
\bm{z}(t + h) = \bm{z}(t) + h \cdot f(\bm{z}(t);\bm{\theta}_f),
\end{align}where $h$, which is usually smaller than 1, is a pre-determined step size of the Euler method.The DOPRI method uses a much more sophisticated method to update $\bm{z}(t + h)$ from $\bm{z}(t)$ and dynamically controls the step size $h$. However, those ODE solvers sometimes incur unexpected numerical instability~\cite{zhuang2020adaptive}. For instance, the DOPRI method sometimes keeps reducing the step size $h$ and eventually, an underflow error is produced. To prevent such unexpected problems, several countermeasures were also proposed.

Instead of the backpropagation method, the adjoint sensitivity method is used to train NODEs for its efficiency and theoretical correctness~\cite{NIPS2018_7892}. After letting $\bm{a}_{\bm{z}}(t) = \frac{d \mathcal{L}}{d \bm{z}(t)}$ for a task-specific loss $\mathcal{L}$, it calculates the gradient of loss w.r.t model parameters with another reverse-mode integral as follows:\begin{align*}\nabla_{\bm{\theta}_f} \mathcal{L} = \frac{d \mathcal{L}}{d \bm{\theta}_f} = -\int_{t_m}^{t_0} \bm{a}_{\bm{z}}(t)^{\mathtt{T}} \frac{\partial f(\bm{z}(t);\bm{\theta}_f)}{\partial \bm{\theta}_f} dt.\end{align*}

$\nabla_{\bm{z}(0)} \mathcal{L}$ can also be calculate in a similar way and we can propagate the gradient backward to layers earlier than the ODE if any. It is worth of mentioning that the space complexity of the adjoint sensitivity method is $\mathcal{O}(1)$ whereas using the backpropagation to train NODEs has a space complexity proportional to the number of DOPRI steps. Their time complexities are similar or the adjoint sensitivity method is slightly more efficient than that of the backpropagation. Therefore, we can train NODEs efficiently.

\subsection{Diffusion Equations}
The diffusion equation is to describe the macroscopic behavior of many micro-particles in Brownian motion, resulting from the random movements and collisions of the particles ~\cite{freedman2012brownian}. In mathematics, it is related to Markov processes, such as random walks, and applied in many other fields, such as materials science, information theory, and biophysics~\cite{shikano2014discrete, dos2017random, plastino2018nonlinear, mendes2015nonlinear}. The essence of the diffusion equation with a graph-based representation of data can be written as follows:
\begin{align}
    \frac{d\bm{h}(t)}{dt} = -\Delta \bm{h}(t) = - k \bm{L}\bm{h}(t),
\end{align}where $\bm{h} \in \mathbb{R}^{|V| \times 1}$ is a vector that contains a value for each node, $k$ is a heat capacity coefficient, $\Delta$ means the discrete Laplacian operator, and $\bm{L}$ is the symmetrically normalized augmented graph Laplacian matrix. Therefore, the change of the values in $\bm{h}$ over time $t$ can be describe by the diffusion equation.

In our case, each node has multi-dimensional values, where the heat equation can be written as follows:
\begin{align}\label{eq:diffusion}
    \frac{d\bm{H}(t)}{dt} = -\Delta \bm{H}(t) = - k \bm{L}\bm{H}(t),
\end{align}where $\bm{H}$ is a matrix, each row of which contains multi-dimensional values for each node.

\begin{figure*}
  \centering
  \includegraphics[width=\textwidth]{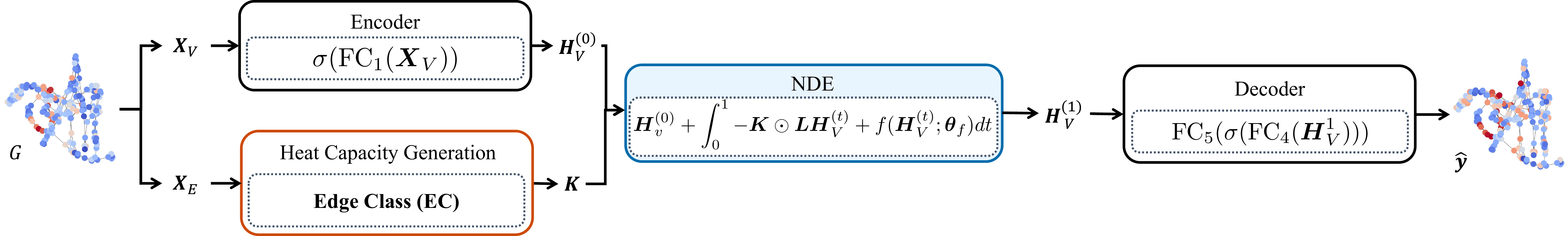}
  \caption{The detailed workflow of NDE}\label{overview}
\end{figure*}

\medskip\noindent{\textbf{Analogy to Simple Graph Convolution (SGC):}} The simple graph convolution~\cite{wu2019simplifying} method is one of the most efficient graph convolutional networks. Its main graph convolutional layer definition is as follows:
\begin{align}\label{eq:sgc}
    \bm{H}(m) = \bm{S}^{m}\bm{H}(0),
\end{align}where $\bm{H}$ is a node hidden matrix, and $\bm{S}^{m}$ is the $m$-th power of the symmetrically normalized adjacency matrix $\bm{S}$ with added self-loops. 

Now we derive Eq.~\eqref{eq:sgc} from Eq.~\eqref{eq:diffusion} to show their analogy~\cite{wang2021dissecting, xhonneux2020continuous}. When applying the Euler discretization to Eq.~\eqref{eq:diffusion} with an interval step size $dt$, we have
\begin{align}\begin{split}
    \bm{H}(t+dt) ={}& \bm{H}(t) - dt \bm{L} \bm{H}(t)\\
    ={}& \bm{H_t} - dt (\bm{I}- \bm{S}) \bm{H}(t) \\
    ={}& [(1-dt)\bm{I}+dt \bm{S} ] \bm{H}(t) \\
    \stackrel{\text{def}}{=}{}& \bm{S}^{dt} \bm{H}(t) \\
\end{split}\end{align}

We will get the following final $\bm{H}(T)$, if we keep evolving the ODE until the terminal time $T$.

\begin{align}
    \bm{H}(T) = [\bm{S}^{dt}]^{m} \bm{H}.
\end{align}

We regard that SGC corresponds to the Euler discretization with a unit step size $dt = 1$. This step size reduces the diffusion matrix to the Linear GCN diffusion matrix $\bm{S}$:

\begin{align}\begin{split}
    \bm{S}^{(dt)} \vert_{dt =1} = (1-1) \bm{I} + \bm{S} = \bm{S}
\end{split}\end{align}

and the final $\bm{H}$ becomes equivalent:

\begin{align}\begin{split}\label{eq:final}
    \bm{H}(T) \vert _{dt =1} = \bm{H}(m) = \bm{H}^{(m)} = \bm{S}^{m} \bm{H}.
\end{split}\end{align}

Therefore, Eq.~\eqref{eq:final} is equivalent to Eq.~\eqref{eq:diffusion} when the Euler method with a unit step size is used. As mentioned earlier, however, we also have many other ODE solvers, such as DOPRI and so forth. Thus, SGC is a special case of the diffusion equation. In this regard, our method, whose one part is the diffusion equation, is able to learn from and infer about graphs. In other words, our proposed model is a spatio-temporal model.

\subsection{Climate Modeling and Weather Forecasting}
Climatologists study the natural factors that cause climate change, using past information to help predict future climate change. Climatological phenomena includes (radiative, convective, and latent) heat transfer, interactions among atmosphere, oceans and surface, and chemical and physical composition of the atmosphere. Climate model elements that can constitute such phenomena are typically differential equations based on physics, fluid motion, and chemistry~\cite{dewan2010understanding}. In weather forecasting tasks, temperature change is related to a transport problem which occurs by diffusion principles. This equation describes a large family of physical processes(heat conduction, wind dynamics, fluid dynamics, etc.)~\cite{hanna1982handbook}. For this reason, we focus on designing our proposed method with the diffusion equation.

Various deep learning-based models are designed for climate and weather forecasting~\cite{ren2021deep,hossain2015forecasting,kurth2018exascale,zaytar2016sequence, rasp2018neural,scher2018toward}, including near-surface air temperature predictions~\cite{seo2019differentiable,seo2019physics}, air quality inference~\cite{cheng2018neural,lin2018exploiting}, precipitation predictions~\cite{zhang2018short, shi2017deep, shi2015convolutional}, wind speed predictions~\cite{cheng2018ensemble, liu2018smart,zhu2018wind} and extreme weather predictions~\cite{racah2016extremeweather, liu2016application }. Climate and weather consists of spatio and temporal data, leading to the development of spatio-temporal models~\cite{tekin2021spatio, chattopadhyay2020predicting}. However, models accounting spatial and temporal dependencies do not consider the differential equations related to the climate model. Recently, researches are trying to leverage physical knowledge for climate predictions. De Bezenac et al. ~\cite{de2019deep} used transport physics (diffusion and advection) to predict the sea surface temperature, but this is limited to a regular grid. DPGN~\cite{seo2019differentiable} designed a physics-informed learning architecture to predict the air temperature. DPGN incorporated differentiable physics equations with a graph neural network framework and used as a physics-informed regularizer. Unlike the above models, we learn the heat capacity of the diffusion equation, which is one important point that have been overlooked for a while, and consider the uncertain nature from the real-word climate data.

\section{Problem Definition}\label{sec:def}
We solve various climate modeling-related problems in this work, using our proposed method. One typical problem is predicting a climate-related factor $a$'s next value from recent values of a set $B$ factors, where i) $a \in B$, e.g., $a$ is next temperature and $B$ includes recent temperature, or ii) $a \notin B$, e.g., $a$ is next temperature and $B$ does not include temperature but humidity, wind velocity, and so on.

It is well known that these processes can be described by diffusion equations under ideal conditions. Due to the uncertain nature of real-world climate data, however, our problem definition is to model the relationships among various noisy climate factors. This problem definition falls into the category of regression.

\section{Proposed Method}
We describe our proposed neural diffusion equation method to model the spatio-temporal relationships among climate factors.

\subsection{Overall Architecture}
Our method consists of four modules as shown in Fig.~\ref{overview}: i) encoding layer, ii) heat capacity generation layer, iii) neural diffusion equation layer, and iv) decoding layer. Given a graph $G=(V,E)$, a node feature matrix $\bm{X}_V \in \mathbb{R}^{|V| \times M}$, and an edge feature matrix $\bm{X}_E \in \mathbb{R}^{|E| \times N}$, the role of each module is as follows:
\begin{enumerate}
    \item The encoding layer converts $\bm{X}_V$ into a hidden matrix $\bm{H}_V^{(0)} \in \mathbb{R}^{|V| \times D}$, where $D$ is the size of hidden vector;
    \item The heat capacity generation layer creates the heat capacity coefficient(s). Our method supports four different heat capacity concepts and the details will be described shortly;
    \item The neural diffusion equation layer processes $\bm{H}_V^{(0)}$ to generate $\bm{H}_V^{(1)}$, following the learned neural diffusion equation;
    \item The decoding layer predicts a target climate factor for each node from $\bm{H}_V^{(1)}$.
\end{enumerate}

The node feature matrix $\bm{X}_V$ can have values for each node for the previous $P$ time steps to learn from historical patterns.

\subsection{Initial Encoding Layer}
The node feature matrix $\bm{X}_V \in \mathbb{R}^{|V| \times M}$ contains the input features of nodes. We use the following encoding layer to produce their initial hidden vectors:
\begin{align}
    \sigma(\mathtt{FC}_1(\bm{X}_V)),
\end{align}where $\sigma$ is a rectified linear unit (ReLU) and $\mathtt{FC}$ is a fully-connected layer.

Instead of directly feeding the raw climate factors contained by $\bm{X}_V$ into the neural diffusion layer, we first create initial hidden vectors.

\subsection{Heat Capacity Generation Layer}\label{sec:hc}
The heat capacity generation layer produces one of the four different types of the heat capacity from the edge feature matrix $\bm{X}_E \in \mathbb{R}^{|E| \times N}$. In diffusion equations, a heat capacity coefficient of an edge represents how quickly (or easily) climate factors are diffused via the edge. Therefore, it is one of the most crucial points in our model to learn a reliable heat capacity. We support the following four types:
\begin{enumerate}
    \item \textbf{Edge Class (EC):} We learn a heat capacity coefficient for each class of edges when edge features are one-hot vectors. Those edges which share the same one-hot vector also share the same heat capacity coefficient, denoted $k_{c} \in \mathbb{R}$, where $c$ means a class of edges;
    \item \textbf{Heat Matrix (HM):} We learn a heat capacity matrix $\bm{K} \in \mathbb{R}^{|V| \times |V|}$. We learn the entire matrix in this case;
    \item \textbf{Single Coefficient (SC):} We learn a single heat capacity coefficient $k \in \mathbb{R}$ which will be shared by all edges;
    \item \textbf{Fixed Coefficient (FC):} We also consider the case of $k=1$. In fact, this is one of the ablation study models.
\end{enumerate}

The first method, denoted \textbf{EC}, can be efficiently implemented using the embedding layer API of PyTorch and TensorFlow, which outputs an embedding vector given a one-hot vector. In our case, it outputs a scalar heat capacity coefficient. 

\subsection{Neural Diffusion Equation Layer}
This neural diffusion equation layer is the core of our proposed method. It evolves $\bm{H}_V^{(0)}$ to $\bm{H}_V^{(1)}$ using the learned neural diffusion equation. The key computation method is as follows:
\begin{align}\label{eq:nde}
    \bm{H}_V^{(1)} = \bm{H}_V^{(0)} + \int_{0}^{1}- \bm{K} \odot \bm{L}\bm{H}_V^{(t)} + f(\bm{H}_V^{(t)},t;\bm{\theta}_f) dt, 
\end{align}where $\odot$ is an element-wise product, and $\bm{L}$ is the symmetrically normalized augmented graph Laplacian matrix of the graph $G$. In the above equation, we assume the heat capacity of \textbf{EC} and \textbf{HM} since in both cases the heat capacity is represented by a matrix $\bm{K}$. For \textbf{SC} and \textbf{FC}, we use a scalar coefficient $k$.

Without the neural network $f$, Eq.~\eqref{eq:nde} reduces to the diffusion equation with the discrete Laplace operator. In real-world environments, however, it is hard to say that diffusion processes are completely governed by the diffusion equation --- in particular, we observe in our experiments that diffusion processes around large cities have non-trivial uncertainties that cannot be solely described by the diffusion equation. To this end, we let the neural network $f$ learn the uncertainty. Therefore, $\frac{d\bm{H}_V^{(t)}}{dt}$ is governed by the heat equation and the learned neural network. The definition of $f$ is as follows:
\begin{align}\label{eq:mininet}
    f(\bm{H}_V^{(t)},t;\bm{\theta}_f) \stackrel{\text{def}}{=} \sigma(\mathtt{FC}_3(\sigma(\mathtt{FC}_2(\bm{H}_V^{(t)})))).
\end{align}

\subsection{Decoding Layer}
The decoding layer is to produce predictions from $\bm{H}_V^{(1)}$. The decoding layer definition is as follows:
\begin{align}
    \hat{\bm{y}} = \mathtt{FC}_5(\sigma(\mathtt{FC}_4(\bm{H}_V^{(1)}))),
\end{align}where $\hat{\bm{y}} \in \mathbb{R}^{|V| \times 1}$ contains a climate factor value for each node.

\subsection{Multi-step Forecasting}
While the default forecasting is to predict very next value for a target climate factor from $\bm{X}_V$, our method also supports predicting for next multiple $S$ time steps. We keep evolving $\bm{H}_V^{(i-1)}$ to $\bm{H}_V^{(i)}$ until $i=S$ as follows, using the neural diffusion equation layer:
\begin{align}\begin{split}
    \bm{H}_V^{(1)} &= \bm{H}_V^{(0)} + \int_{0}^{1}- \bm{K} \odot\bm{L}\bm{H}_V^{(t)} + f(\bm{H}_V^{(t)},t;\bm{\theta}_f) dt,\\
    \bm{H}_V^{(2)} &= \bm{H}_V^{(1)} + \int_{1}^{2}- \bm{K} \odot\bm{L}\bm{H}_V^{(t)} + f(\bm{H}_V^{(t)},t;\bm{\theta}_f) dt,\\
    &\vdots\\
    \bm{H}_V^{(S)} &= \bm{H}_V^{(S-1)} + \int_{S-1}^{S}- \bm{K}\odot \bm{L}\bm{H}_V^{(t)} + f(\bm{H}_V^{(t)},t;\bm{\theta}_f) dt.
\end{split}\end{align}

One can consider that this mechanism corresponds to a continuous recurrent network~\cite{debrouwer2019gruodebayes, kidger2020neural}. The decoding layer then predicts for each time step $s \in \{1,2,\cdots,S\}$ using $\bm{H}_V^{(s)}$.

\subsection{Training Algorithm}
\begin{algorithm}[t]
\small
\SetAlgoLined
\caption{How to train our proposed NDE}\label{alg:train}
\KwIn{Training data $D_{train}$, Validating data $D_{val}$, Maximum iteration number $max\_iter$}
Initialize $\bm{\theta}_f$ and other parameters $\bm{\theta}_{others}$ if any, e.g., the parameters of the encoding, decoding, and other layers;

$k \gets 0$;

\While {$k < max\_iter$}{
    Train $\bm{\theta}_f$ and $\bm{\theta}_{others}$\;\label{alg:train1}
    
    Validate and update the best parameters, $\bm{\theta}^*_f$ and $\bm{\theta}^*_{others}$, with $D_{val}$\;
    
    $k \gets k + 1$;
}
\Return $\bm{\theta}^*_f$ and $\bm{\theta}^*_{others}$;
\end{algorithm}

We use Alg.~\ref{alg:train} to train our proposed NDE. For the gradient calculation, we use the following mean squared error loss function:
\begin{align}
    \mathcal{L} \stackrel{\text{def}}{=} \frac{\sum_{i=1}^{|V|} (\bm{y}_{[i]} - \hat{\bm{y}}_{[i]})^2}{|V|},
\end{align}where $\bm{y}_{[i]}$ means the $i$-th element of $\bm{y}$.

The training algorithm follows a standard method to update the parameters and validate with a validation set. One more thing worth mentioning is that the training process can be theoretically well-posed.

\begin{wrapfigure}{l}{0.4\columnwidth}
\centering
\includegraphics[width=0.4\columnwidth]{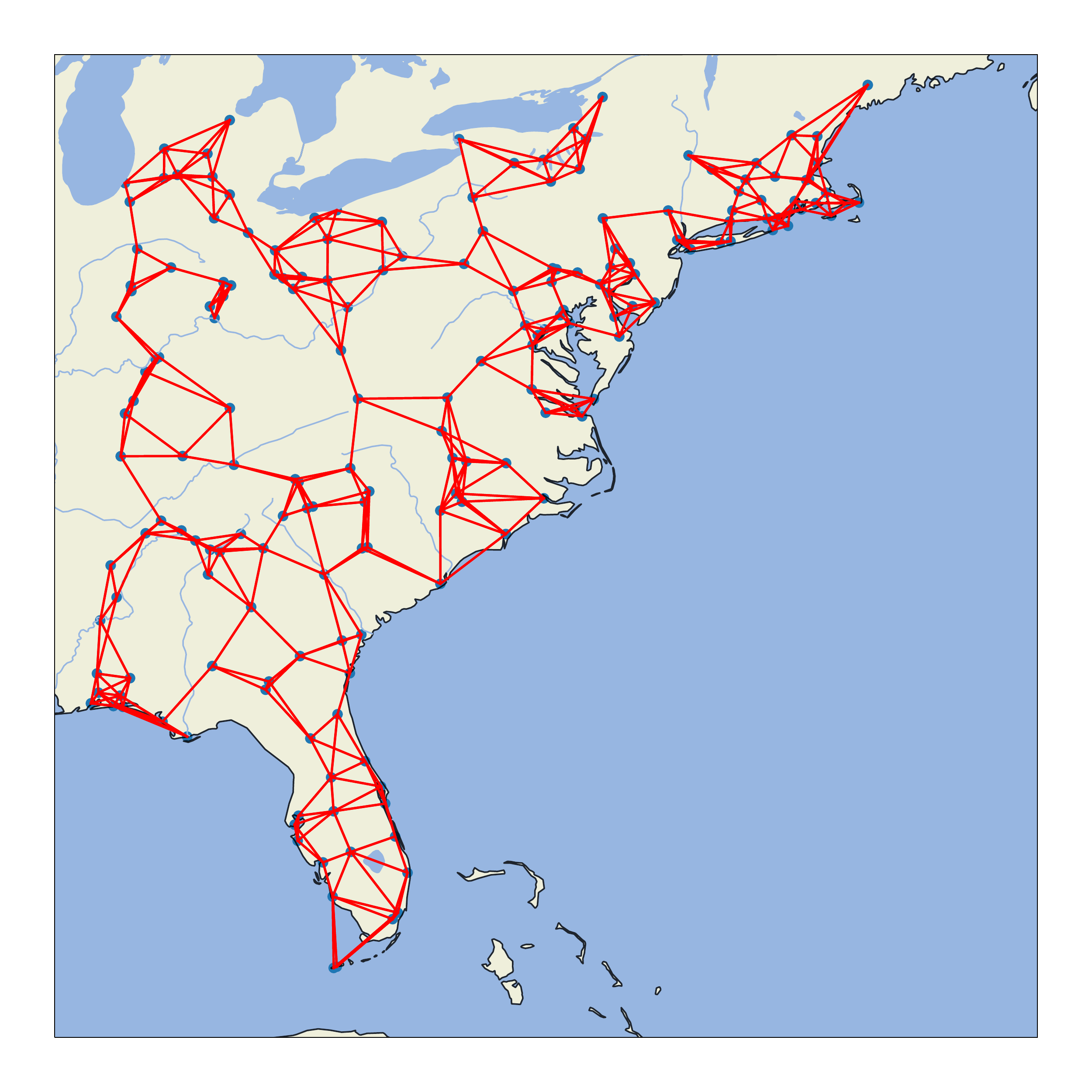}
\caption{Weather stations in the eastern states of the US and its $4$-NN graph. } \label{fig:noaa}
\end{wrapfigure}

\medskip\noindent{\textbf{Well-posedness of the Solution of NDE:}} The well-posedness\footnote{A well-posed problem means i) its solution uniquely exists, and ii) its solution continuously changes as input data changes.} of ODEs was already proved in \cite[Theorem 1.3]{lyons2004differential} under the mild condition of the Lipschitz continuity of $\frac{d\bm{h}(t)}{dt}$, i.e., $\frac{d\bm{H}(t)}{dt}$ in our case. The matrix multiplication of the heat equation is a representative Lipschitz continuous operator with a specific Lipschitz constant. In addition, almost all activations, such as ReLU, Leaky ReLU, SoftPlus, Tanh, Sigmoid, ArcTan, and Softsign, have a Lipschitz constant of 1. Other common neural network layers, such as dropout, batch normalization and other pooling methods, have explicit Lipschitz constant values. Therefore, the Lipschitz continuity of $\frac{d\bm{H}(t)}{dt}$ can be fulfilled in our case, and our training algorithm solves a well-posed problem so its training process is stable in practice.

\section{Experiments}
We conduct experiments with synthetic and real-world datasets for climate modeling. All experiments were conducted in the following software and hardware environments: \textsc{Ubuntu} 18.04 LTS, \textsc{Python} 3.8.0, \textsc{Numpy} 1.18.5, \textsc{Scipy} 1.5, \textsc{Matplotlib} 3.3.1, \textsc{PyTorch} 1.7.0, \textsc{CUDA} 10.0, and \textsc{NVIDIA} Driver 417.22, i9 CPU, and \textsc{NVIDIA RTX Titan}. We repeat the training and testing procedures with ten different random seeds and report their mean and standard deviation accuracy --- resources are accessible at {\color{blue}\url{https://github.com/jeehyunHwang/Neural-Diffusion-Equation}}.

\begin{table}[t]
\centering
\setlength{\tabcolsep}{2pt}
\caption{Dataset description}\label{tbl:dataset}
\begin{tabular}{ccc}
\hline
Dataset & LA \& SD & NOAA \\ \hline
Train & 2012/6/28 9pm-2012/07/8 7am & 2015/1/1 0am-2015/9/13 12pm\\
Valid & 2012/7/8 7am-2012/07/10 9am & 2015/9/13 12pm-2015/10/20 0am\\
Test & 2012/7/10 9am-2012/07/14 10pm & 2015/10/20 0am-2016/1/1 0am \\
\hline
\end{tabular}
\end{table}

\begin{figure}[!t]
\centering
\subfigure[LA ($t=1$)]{\includegraphics[width=0.24\columnwidth]{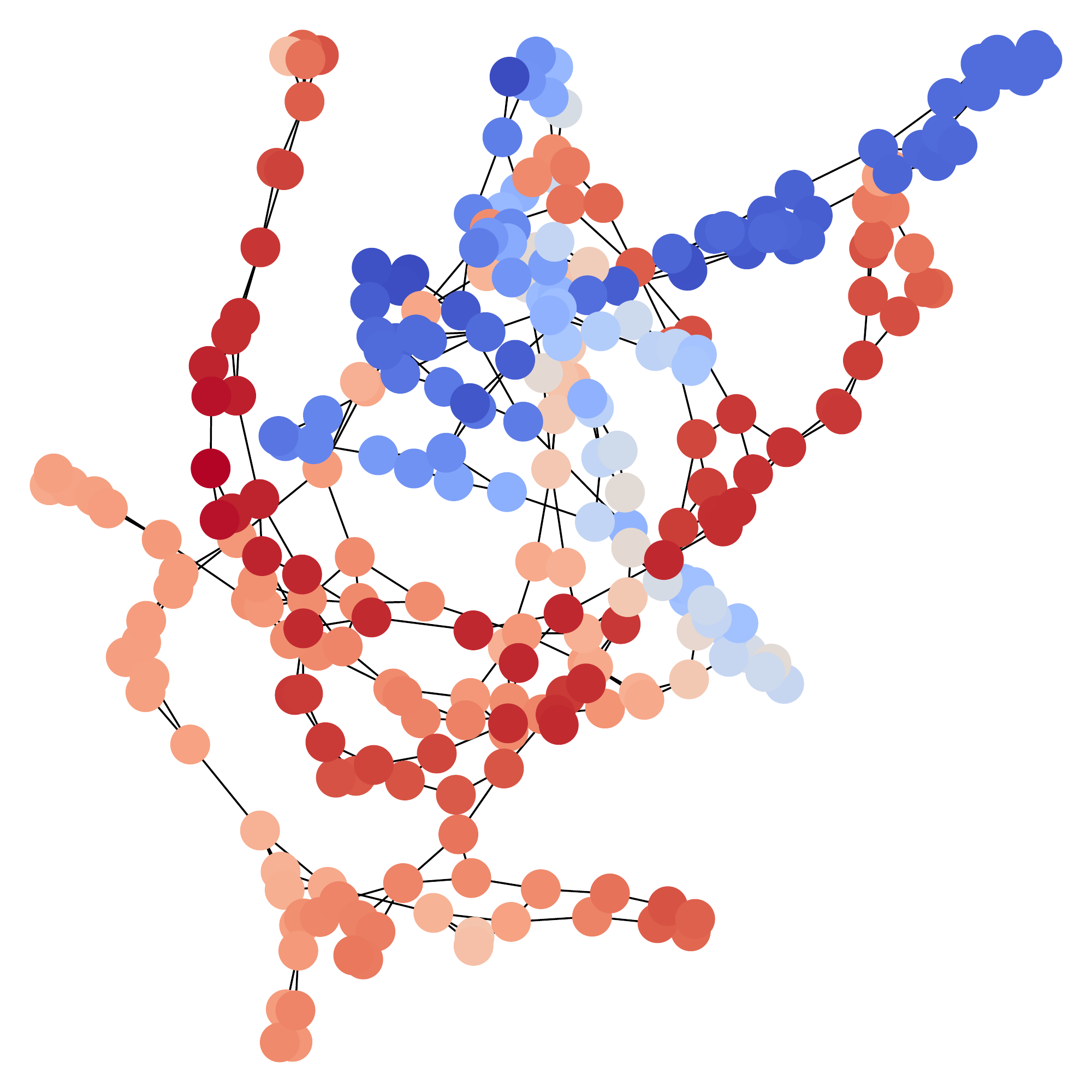}}
\subfigure[LA ($t=2$)]{\includegraphics[width=0.24\columnwidth]{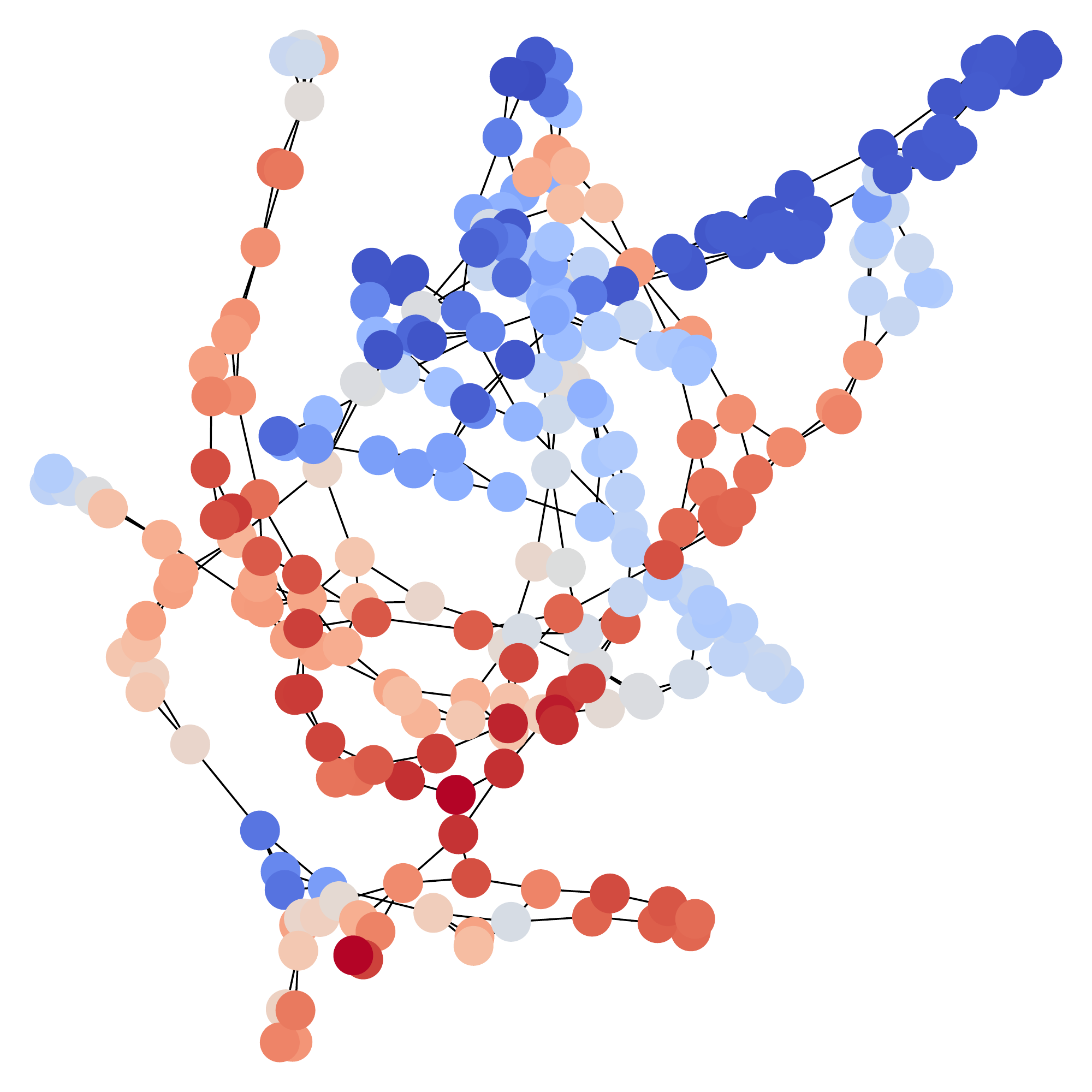}}
\subfigure[LA ($t=3$)]{\includegraphics[width=0.24\columnwidth]{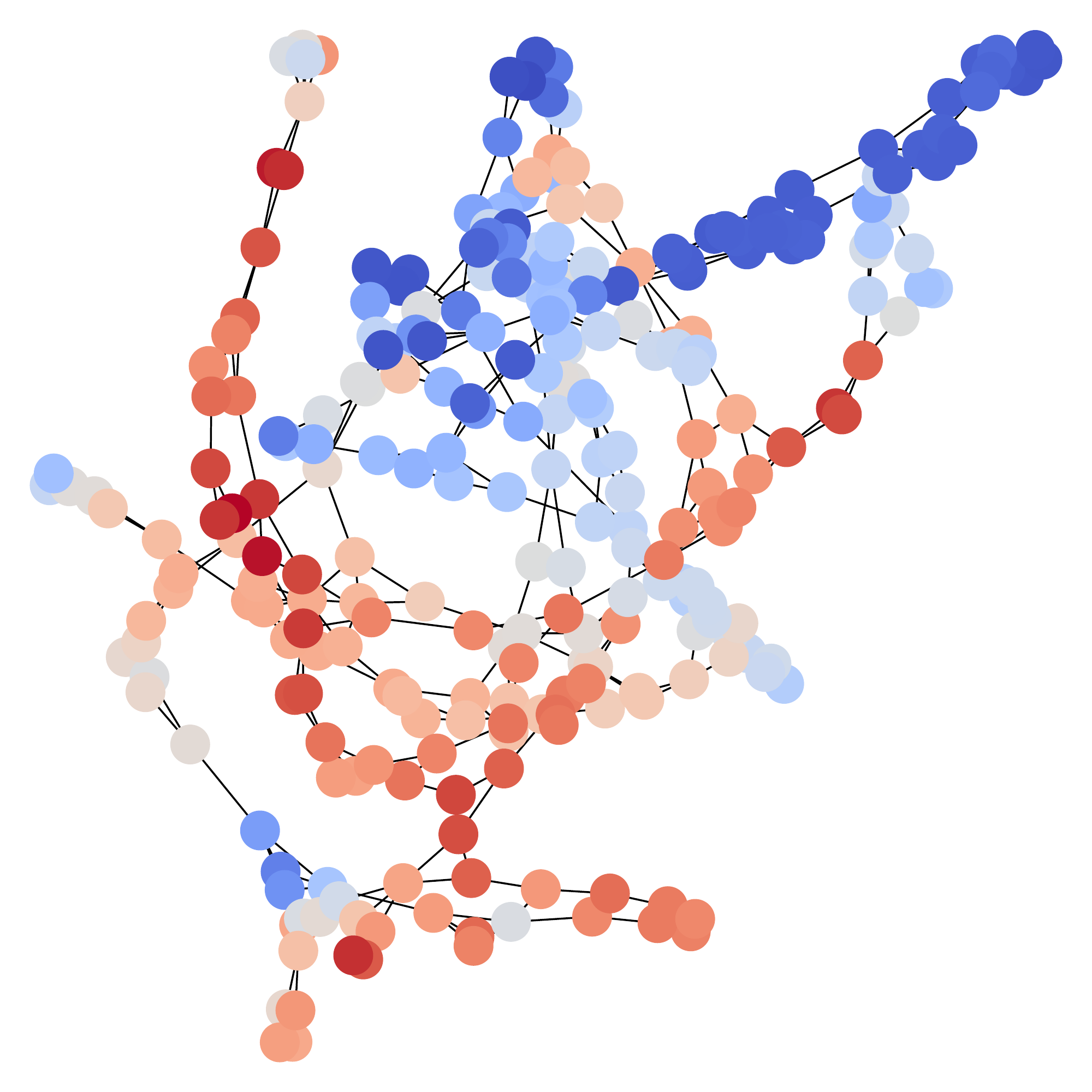}}
\subfigure[LA ($t=4$)]{\includegraphics[width=0.24\columnwidth]{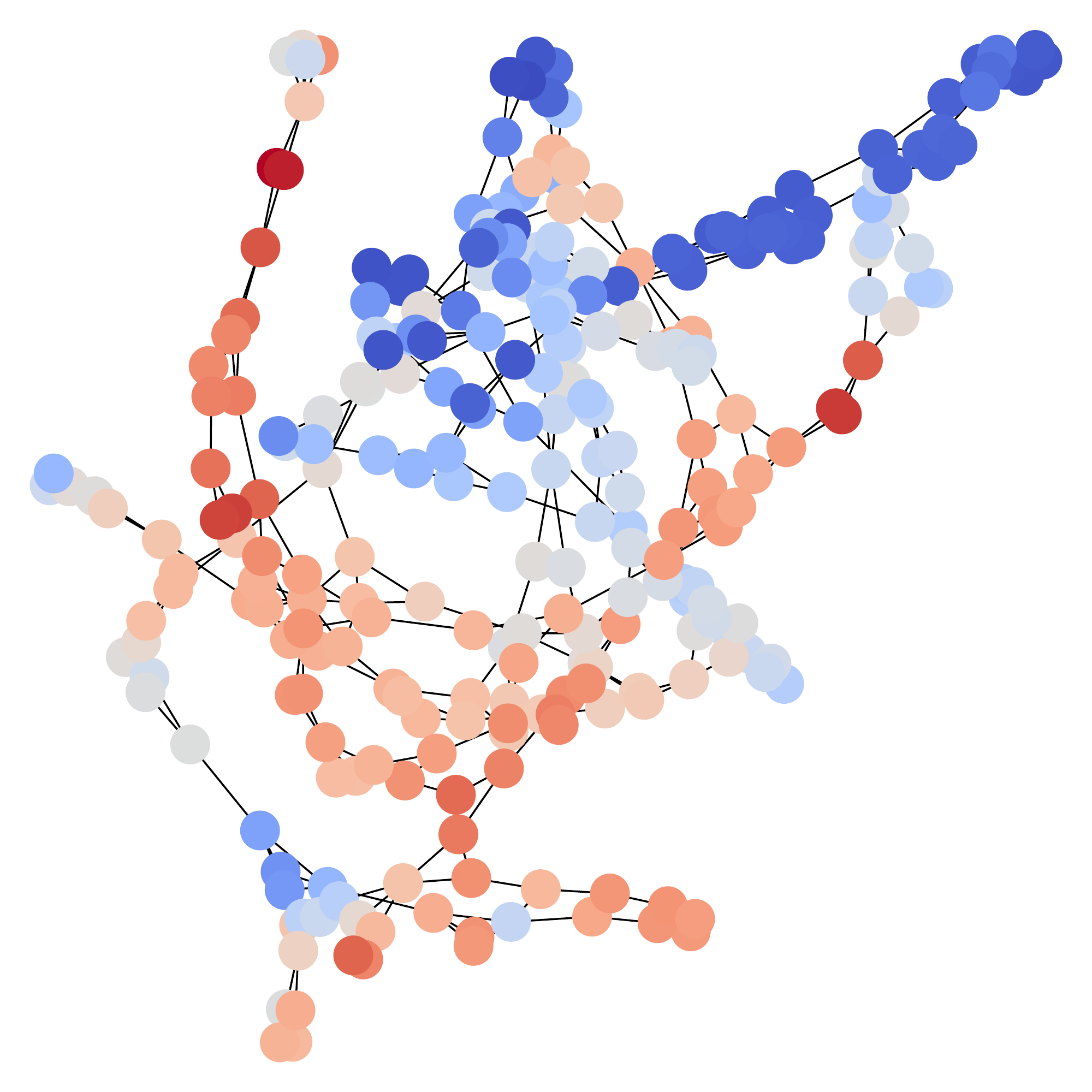}}
\subfigure[SD ($t=1$)]{\includegraphics[width=0.24\columnwidth]{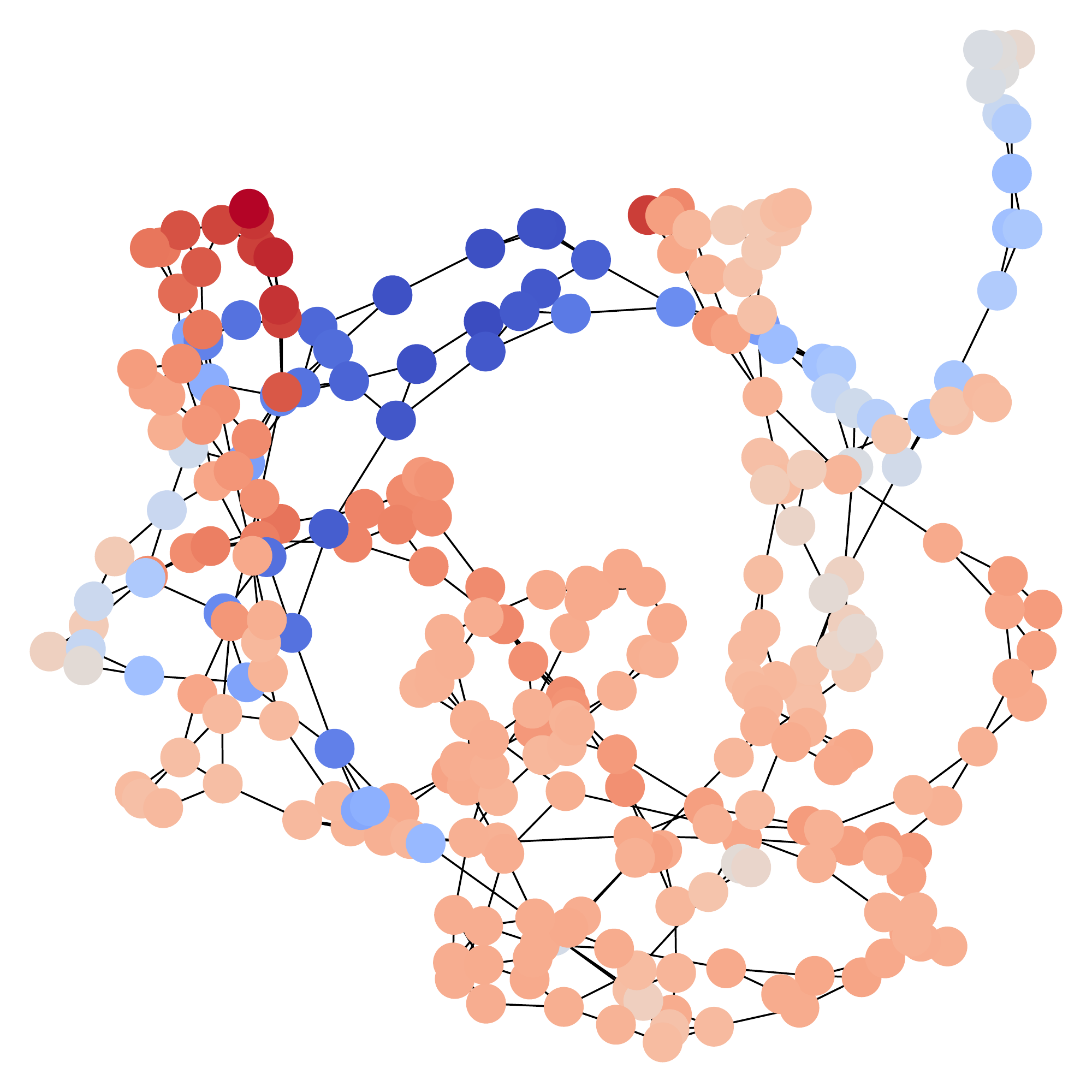}}
\subfigure[SD ($t=2$)]{\includegraphics[width=0.24\columnwidth]{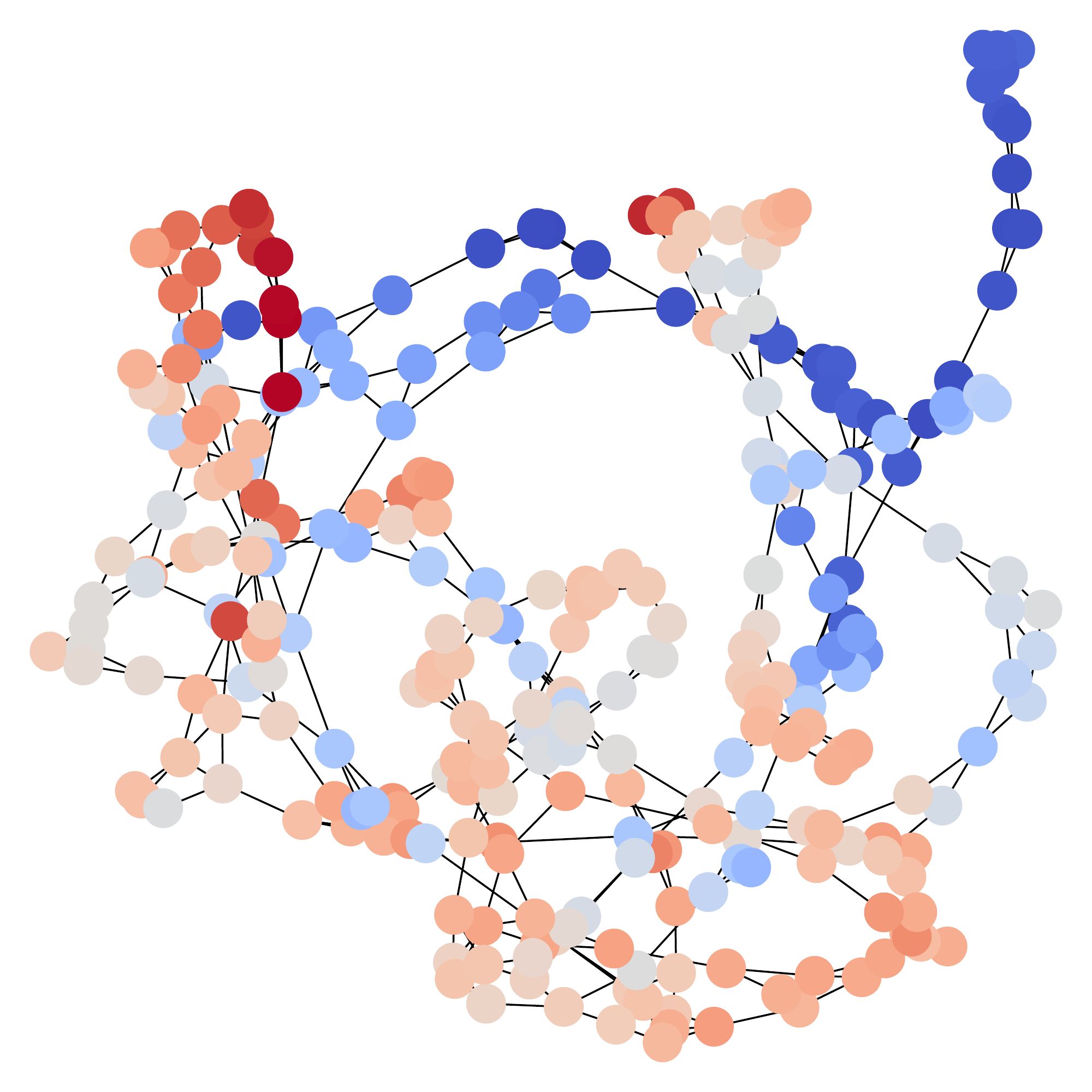}}
\subfigure[SD ($t=3$)]{\includegraphics[width=0.24\columnwidth]{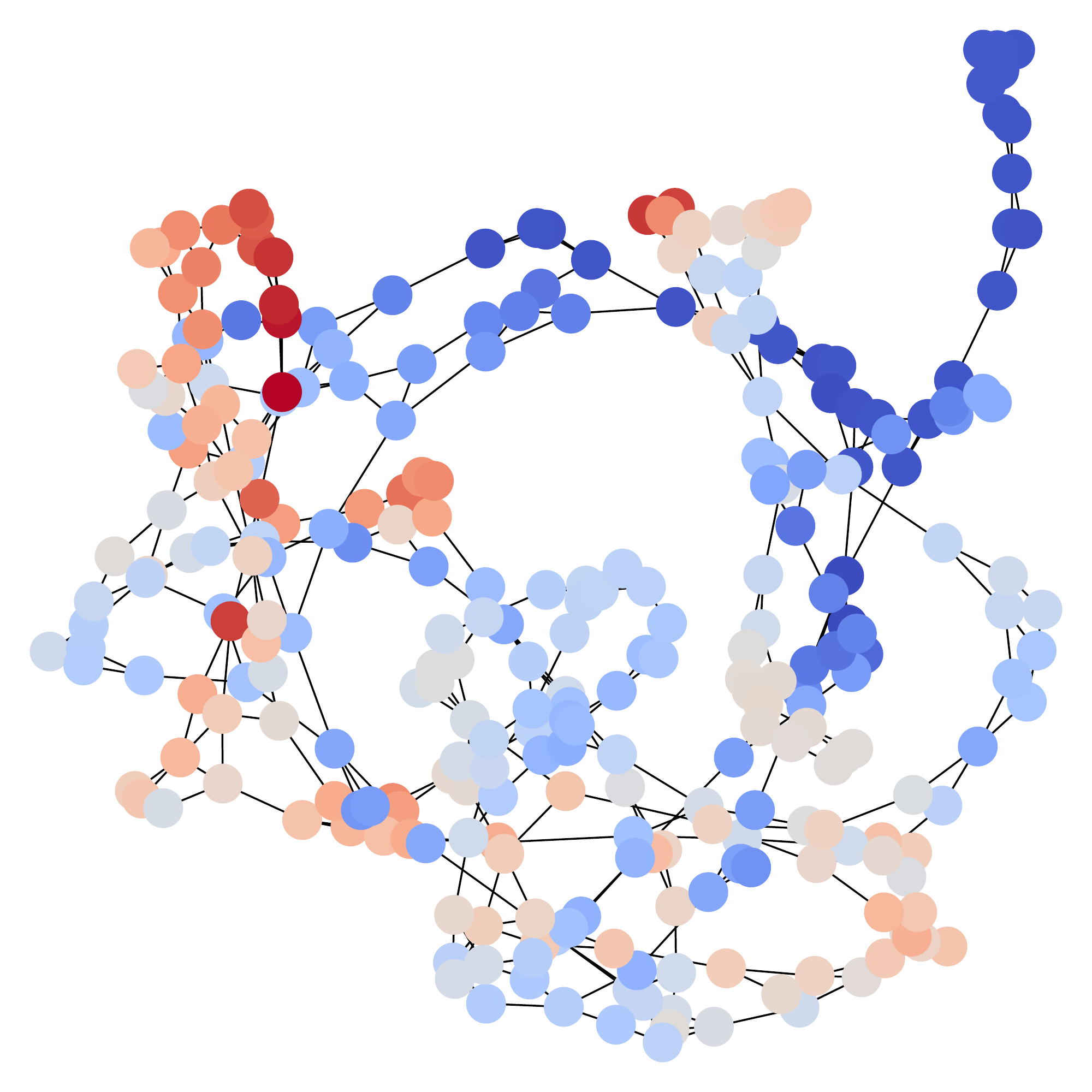}}
\subfigure[SD ($t=4$)]{\includegraphics[width=0.24\columnwidth]{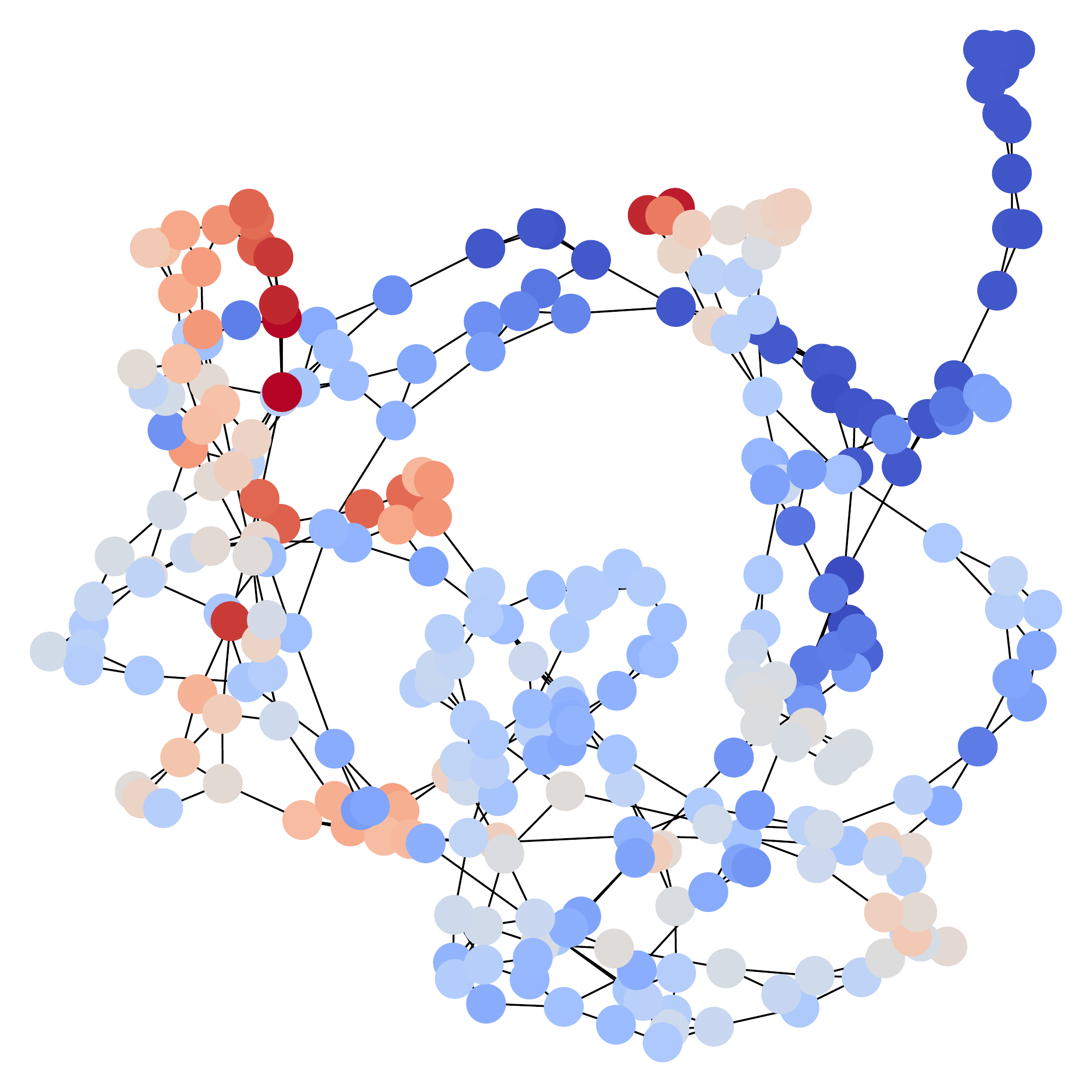}}
\caption{The graph representation of the LA and SD areas. There are 274 and 282 nodes, respectively. These graphs show the change of the air temperature over time.}\label{fig:LA&SD}
\end{figure}

\subsection{Experimental Environments}

\subsubsection{Synthetic Data}
We test with synthetic data from physics. Network dynamics are generated based on the analytical solutions of the diffusion equation referred in Eq.~\eqref{eq:diffusion}. We consider a grid network and a small-world network. Each node of the grid network has eight connected neighbors. In particular, we add the small-world network due to the fact that climate networks can be regarded as small-world networks~\cite{gozolchiani2011emergence,yang2001small}. In fact, the real-world datasets we use later are classified as small-world networks because all the small-coefficients, which are coefficients for quantifying small-worldness~\cite{humphries2006brainstem, humphries2008network}, are greater than 1. We use the Watts-Strogats model to generate the small-world network~\cite{watts1998collective}. We first generate ground-truth values of each network dynamics including 400 nodes using the Dormand-Prince method~\cite{dormand1996numerical}. We sample 100 snapshots of the created continuous-time dynamics with the same time-interval, and use first 80 snapshots for training and the remaining 20 snapshots for testing. 

\subsubsection{LA and SD Data}
We also use a dataset that consists of hourly climate observations over 16 days (June/28/2012 21:00 to July/14/2012 22:00) in Los Angeles and San Diego areas, including 274 and 282 nodes, respectively (as shown in Fig.~\ref{fig:LA&SD}). Each node includes 10 climate observations: air temperature, albedo, precipitation, soil moisture, relative humidity, specific humidity, surface pressure, planetary boundary layer height, and wind vector (2 directions). Listed observations are selected due to its characteristic of high relation to diffusion equation~\cite{stocker2011introduction}. The dataset also includes edge attributes representing the land-usage types of source and destination nodes, resulting in 43 different edge connection classes. The discrete time unit is one hour. The statistics of all real-world datasets we use for our experiments are summarized in Table~\ref{tbl:dataset}.

\subsubsection{NOAA Data}
We conduct experiments with a dataset including hourly average temperature derived from the Online Climate Data Directory of the National Oceanic and Atmospheric Administration(NOAA)\footnote{https://www.ncdc.noaa.gov/cdo-web/}. We select 188 stations located in a region of the latitude in [23.886, 44.371] and the longitude in [-87.188, -67.605] as shown in Fig.~\ref{fig:noaa}, from 2015 of hourly climate normals dataset. The imputation of missing values is done using the average value of $2$ time-wise nearest neighbors for experimental purposes. We adopt $4$-NN (Nearest Neighbor) algorithm to construct graph structure, and the created adjacency matrix $\bm{A}$ is converted by $\bm{A}=(\bm{A}+\bm{A}^{T})/2$ to make it symmetric. The discrete time unit is one hour.

\subsubsection{Evaluation Method}
For the synthetic dataset, we perform a time-series forecasting of reading the first 80\% and predicting the last 20\% of observations, i.e., extrapolation. We exclude the interpolation from this evaluation since almost all models show reasonable accuracy for it. However, they show significantly different accuracy for the extrapolation. Since each node has one scalar value strictly following the diffusion equation, this dataset is relatively more straightforward than other real-world datasets.

With the LA and SD datasets, we feed $P$ recent observations, where $P=\{1,3,5\}$, and predict $S$ next values, where $S=\{1,10\}$, i.e., sequence-to-sequence predictions. This specific experimental setting had been used in~\cite{seo2019differentiable} as well. In particular, the input node feature does not include the air temperature but the ground-truth output is the air temperature, i.e., the case of $a \notin B$ in Section.~\ref{sec:def}. This forecasting can be used to reconstruct the sensing values of malfunctioning/dead sensors from others.

In the NOAA dataset, we predict next $S$ temperature values, where $S=\{1,6\}$, from recent $P$ temperature values, where $P=\{1,6\}$, i.e., the case of $a \in B$ in Section.~\ref{sec:def}.

In all datasets, we use the evaluation metrics of mean absolute error (MAE) and/or mean squared error (MSE). In all cases, we run with 10 different random seeds and report their mean and standard deviation accuracy.

\subsubsection{Baselines}
We consider the following baseline to compare with our method:
\begin{enumerate}
    \item In the first group of baselines, they ignore graph connectivity and perform individual forecasting for each node as follows:
    \begin{enumerate}
        \item Multi-Layer Perceptron (MLP) is a basic neural network model which uses a series of fully-connected layers. We set the number of layers to 2, and the hidden size to 64.
        \item Recurrent Neural Network (RNN), Long Short-Term Memory (LSTM), and Gated Recurrent Units (GRU) are popular models to deal with time-series data. We set the number of layers to 2, and the hidden size to 64.
    \end{enumerate}
    \item The next group includes the temporal-GNN models, which are combinations of GNN and RNN models. The temporal-GNN model adopt the GNN to capture the graph structure and RNN/LSTM/GRU to learn the temporal information.
    \item DPGN~\cite{seo2019differentiable} and NDCN~\cite{zang2020neural} are two differential equation-based models to predict values that follow the diffusion equation. They do not consider learning the heat capacity coefficients and modeling uncertainty.
    \begin{enumerate}
        \item The key of DPGN is injecting the knowledge of the diffusion equation into neural networks. To this end, it uses a technique known as \emph{physics-informed neural network} (PINN~\cite{raissi2019physics}).
    \end{enumerate}
    \item We also consider two variants of DPGN: GN-skip and GN-only. The GN-only model includes three modules, which are a graph encoder, a graph network (GN) block~\cite{sanchez2018graph}, and a graph decoder. The GN-skip model connects the input and output of the GN block with a skip-connection~\cite{he2016deep}. For these two variants of DPGN, however, we do not inject the knowledge of the heat equation.
\end{enumerate}

\subsubsection{Hyperparameters}
We consider the following ranges of the hyperparameters: the hidden vector size $D$ is in \{8, 20, 16, 32\}. The number of layers of $f$ in Eq.~\eqref{eq:mininet} is in \{2,3\}. The hidden vector size $D$ is in \{128, 256\}. The learning rate in all methods is in \{\num{1e-2}, \num{1e-3}, \num{1e-4}\}.

For reproducibility, we report the best hyperparameters for our method as follows:
    \begin{enumerate}
        \item Synthetic Data: the learning rate is \num{1e-2}, the weight decay is \num{1e-3}, the hidden vector size $D$ is 20;
        \item LA: the learning rate is \num{1e-3}, the weight decay is \num{5e-4}, the hidden vector size $D$ is 32;
        \item SD: the learning rate is \num{1e-3}, the weight decay is \num{5e-4}, the hidden vector size $D$ is 32;
        \item NOAA: the learning rate is \num{1e-3}, the weight decay is \num{0}, the hidden vector size $D$ is 32.
    \end{enumerate}

\subsection{Experimental Results}

\subsubsection{Synthetic Data}
Table~\ref{tbl:synthetic} summarizes the mean and standard deviation accuracy of the extrapolation experiments with the synthetic datasets for various models. To inject some uncertainty into the data, we add temporal noises and consider both with and without the noises. RNN-based models, such as RNN and RNN-GNN, show relatively poor accuracy. This is because RNN models do not have enough capacity to learn from the first 80\% of observations, which is rather long to be processed by RNNs. LSTM and GRU-based models show reasonable accuracy. GRU-based models are better than LSTM-based models in our experiments without noises in the grid network. When there exist noises, LSTM outperforms GRU. Both LSTM and GRU are capable of learning from the long sequence, i.e., the first 80\% of observations, and show better accuracy than RNN. One more result worth mentioning is that GRU-GNN outperforms GRU in the grid network when there are no noises, which shows the efficacy of processing both temporal and spatial information.

However, all the best outcomes are made by differential equation-based models, NDCN and NDE. Among them, our NDE shows much smaller MAE values. The MAE of NDE is 55\% of that of NDCN, which is about 45\% smaller for the grid network. In the case of the small-world network, NDE shows a 15\% smaller MAE than that of NDCN. Since our NDE explicitly models noise (uncertainty), it shows better accuracy than others.

\begin{table*}[t]
\centering
\setlength{\tabcolsep}{4pt}
\caption{Prediction errors (MAEs) in synthetic data from heat diffusion. Each result is the mean and the standard deviation with 10 runs.}\label{tbl:synthetic}
\begin{tabular}{cccccc}
\hline
\multirow{2}{*}{Model} & Grid & Small-world & Grid & Small-world  & \multirow{2}{*}{\#Params}\\
    & without noise  & without noise  & with noise & with noise &  \\ \hline
RNN & 0.7188 $\pm$ 0.2464  & 0.1452 $\pm$ 0.0697  & 1.0087 $\pm$ 0.0014 & 0.6423 $\pm$ 0.0590 & 24,530\\
LSTM & 0.5374 $\pm$ 0.1352  & 0.1561 $\pm$ 0.0162 & 0.8230 $\pm$ 0.0531 & 0.5984 $\pm$ 0.0017 & 84,890\\
GRU & 0.4887 $\pm$ 0.0145   & 0.1050 $\pm$ 0.0179 & 0.8815 $\pm$ 0.1021 & 0.6121 $\pm$ 0.0149 & 64,770\\
RNN-GNN & 0.7511 $\pm$ 0.2253 & 0.1783 $\pm$ 0.0641 & 1.0088 $\pm$ 0.0017 & 0.6426 $\pm$ 0.0585 & 24,530\\
LSTM-GNN & 0.5303 $\pm$ 0.0984 & 0.1739 $\pm$ 0.0524 & 0.8182 $\pm$ 0.1016 & 0.6287 $\pm$ 0.0214 & 84,890\\
GRU-GNN & 0.4530 $\pm$ 0.0774 & 0.1054 $\pm$ 0.0178 &1.0042 $\pm$ 0.0692 & 0.6249 $\pm$ 0.0247 & 64,770\\
NDCN & 0.2007 $\pm$ 0.3963 & 0.0799 $\pm$ 0.0194 & 0.6882 $\pm$ 0.0641 &0.5921 $\pm$ 0.2421 & 901\\
\textbf{NDE} & \textbf{0.1121 $\pm$ 0.0229} & \textbf{0.0687 $\pm$ 0.0260} & \textbf{0.4569 $\pm$ 0.0347} & \textbf{0.4822 $\pm$ 0.2124} & 1,761\\
\hline
\end{tabular}
\end{table*}

\begin{figure}[h]
\centering
\subfigure[True ($t=21.5$)]{\includegraphics[width=0.32\columnwidth]{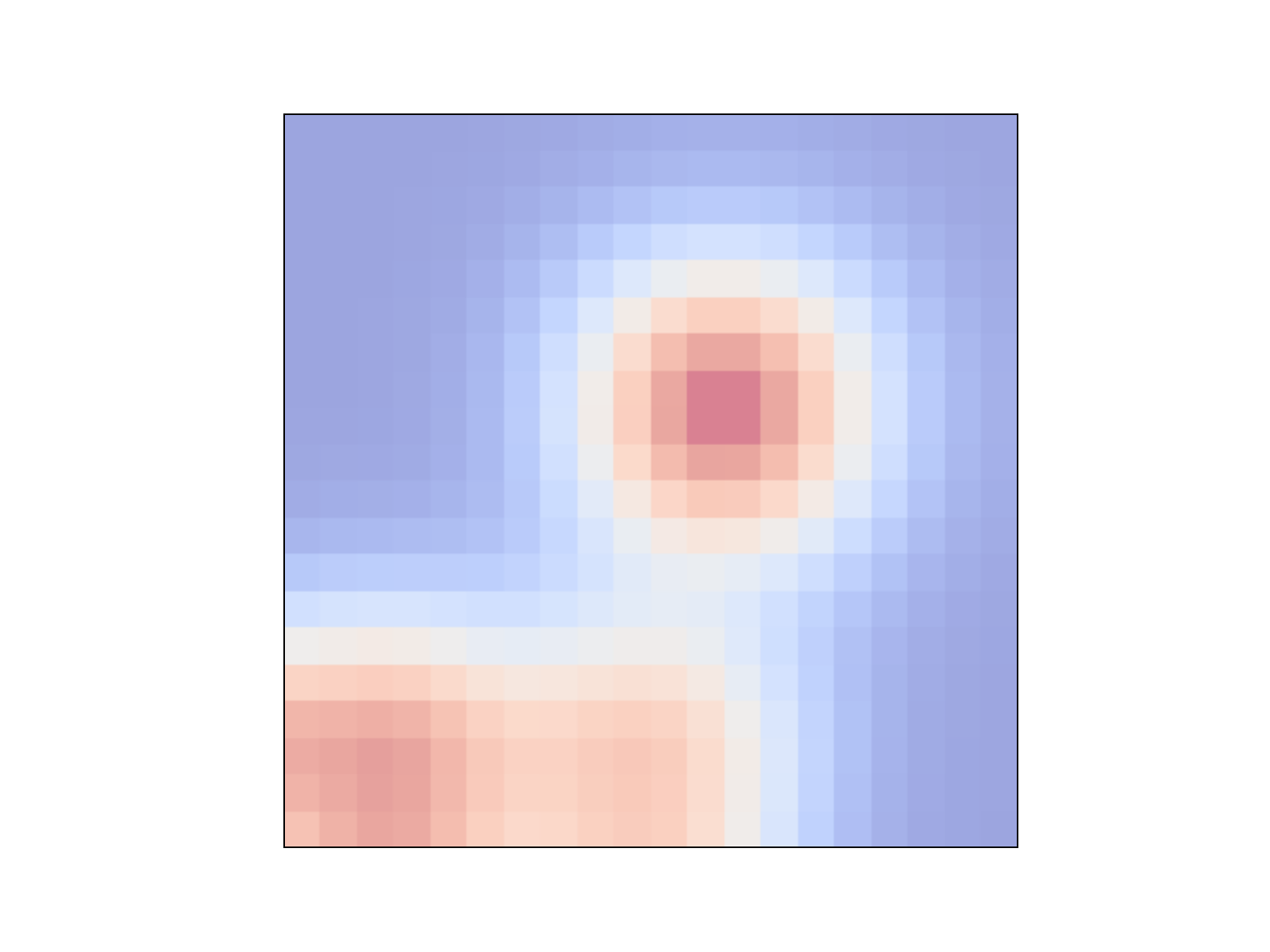}}
\subfigure[True ($t=81$)]{\includegraphics[width=0.32\columnwidth]{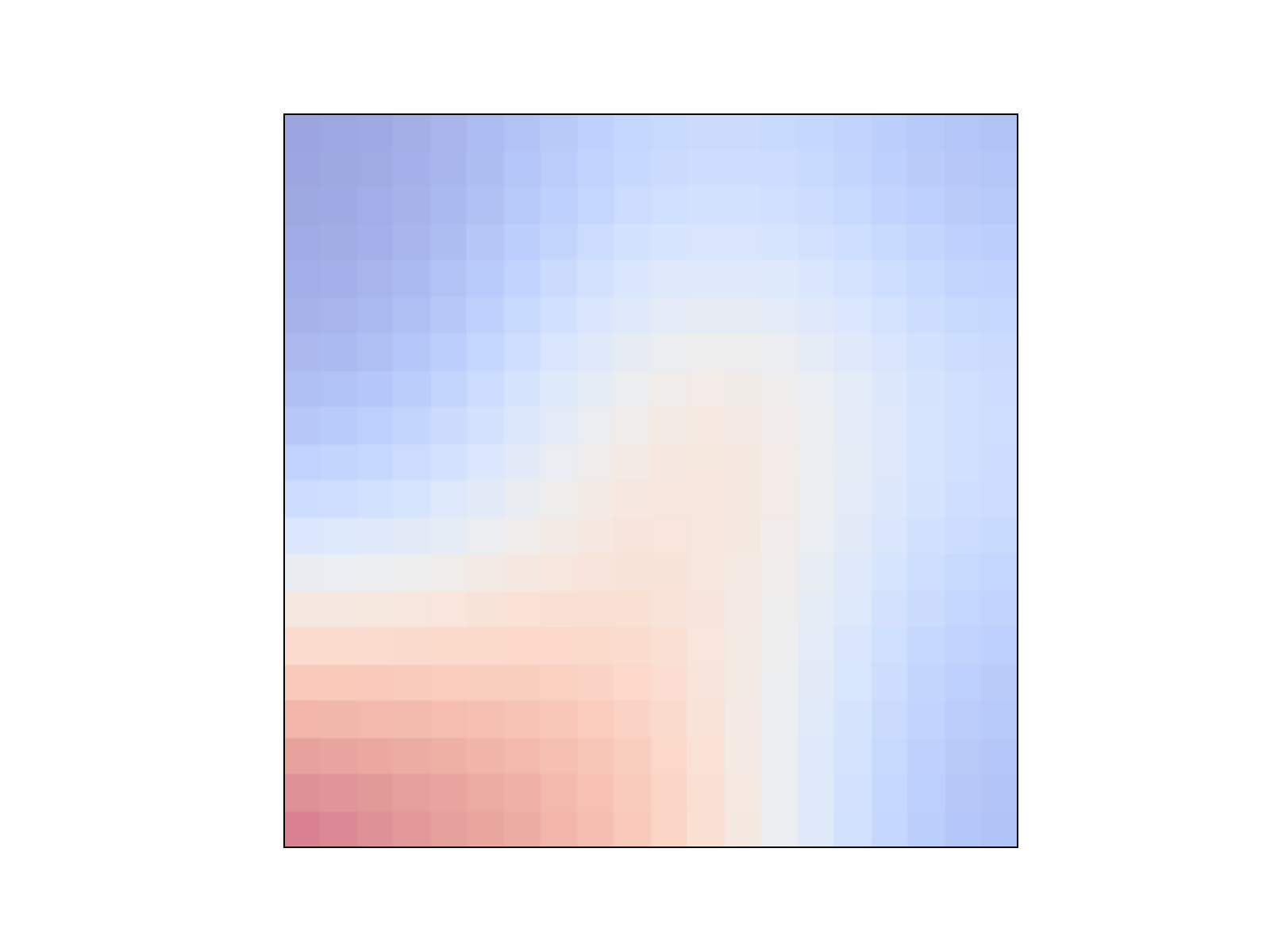}}
\subfigure[True ($t=90$)]{\includegraphics[width=0.32\columnwidth]{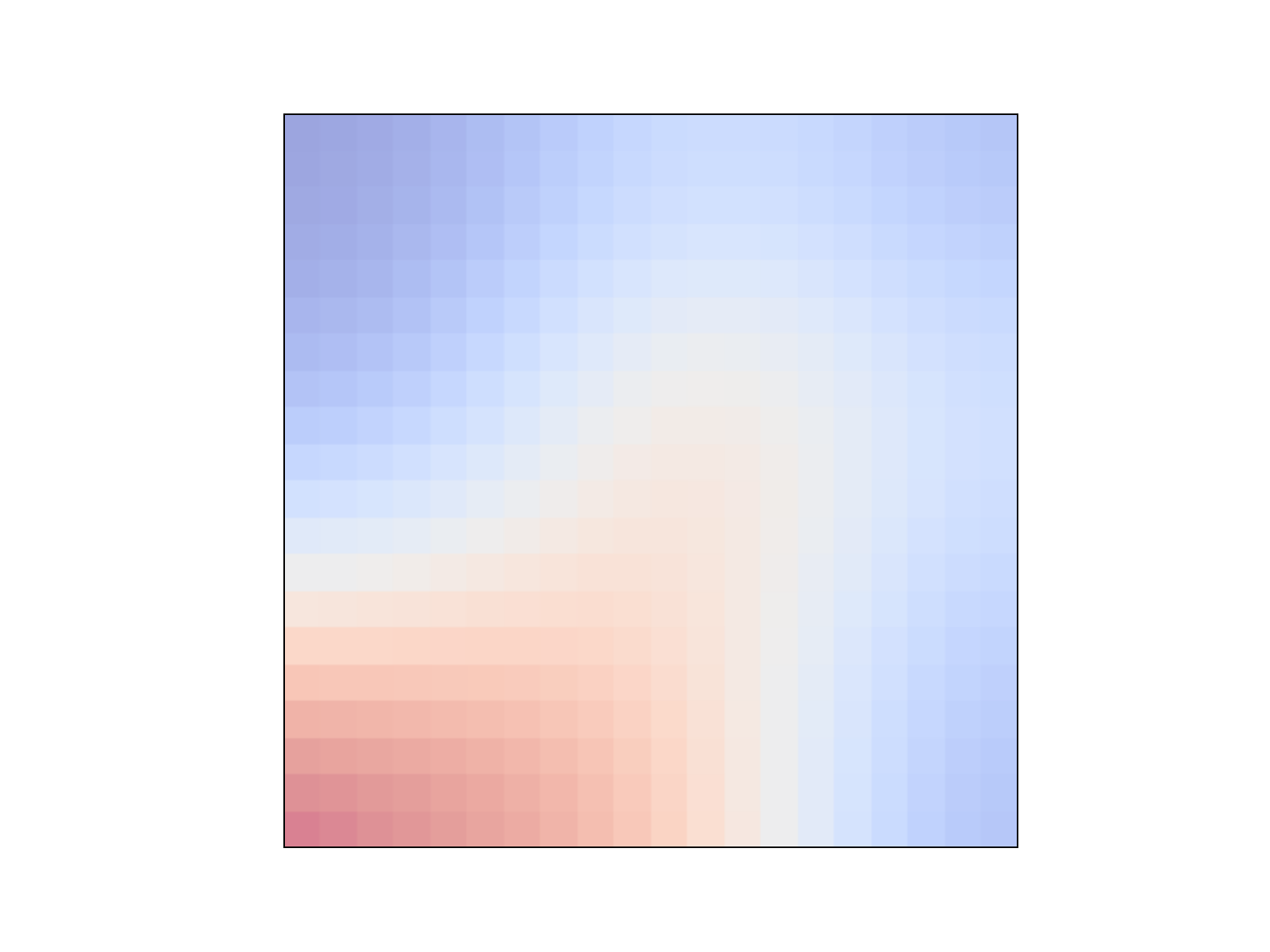}}
\subfigure[NDE ($t=21.5$)]{\includegraphics[width=0.32\columnwidth]{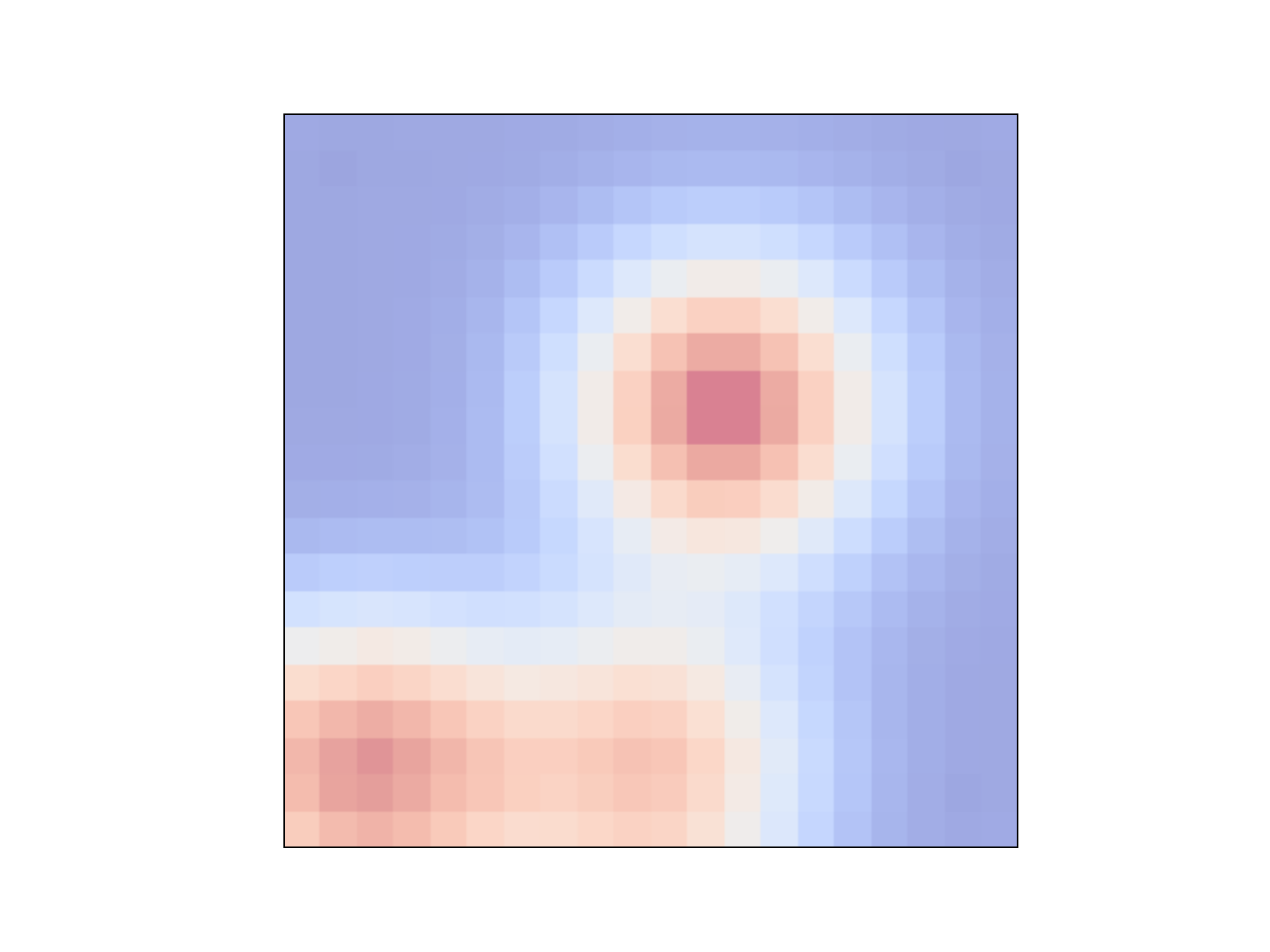}}
\subfigure[NDE ($t=81$)]{\includegraphics[width=0.32\columnwidth]{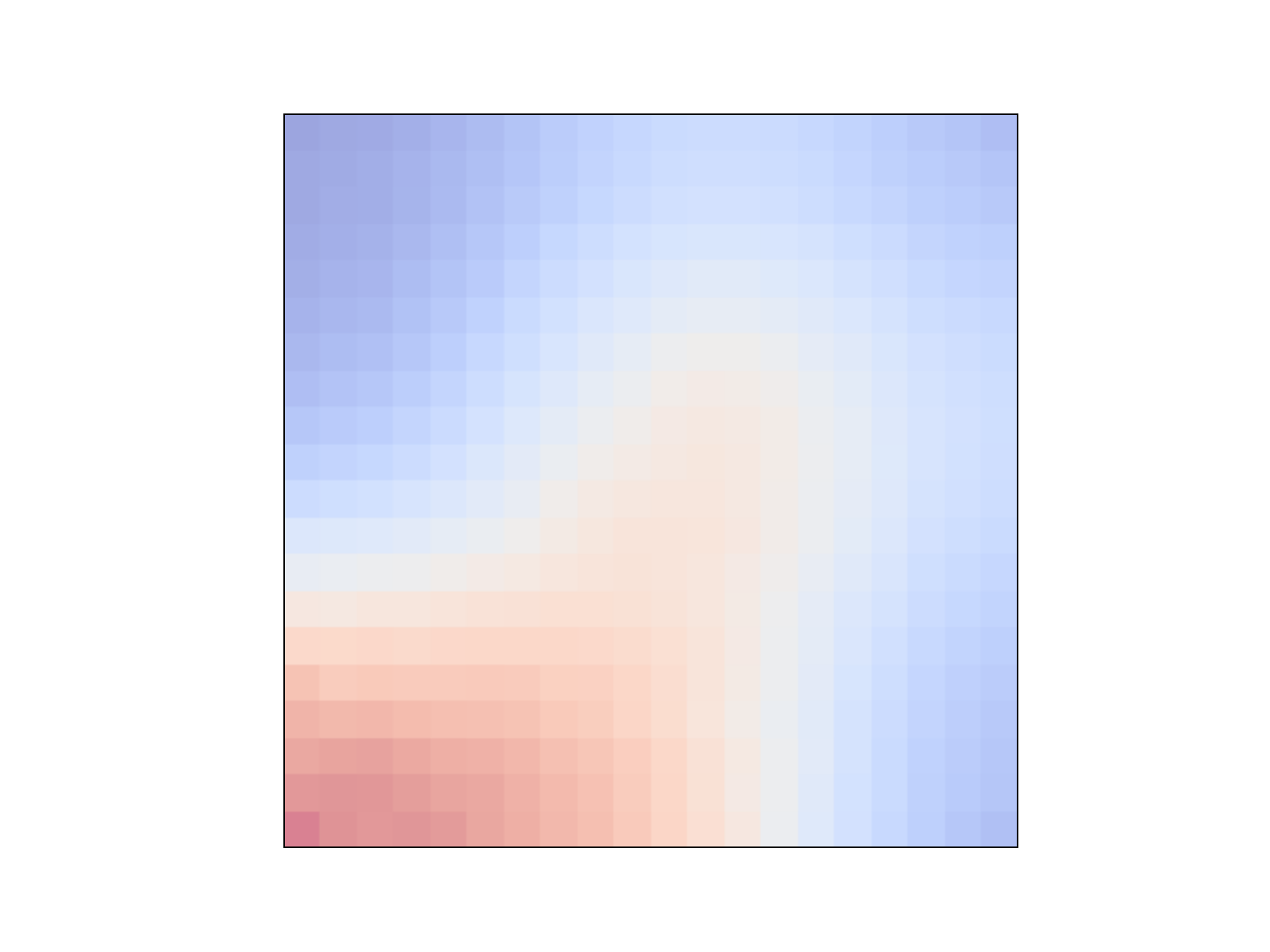}}
\subfigure[NDE ($t=90$)]{\includegraphics[width=0.32\columnwidth]{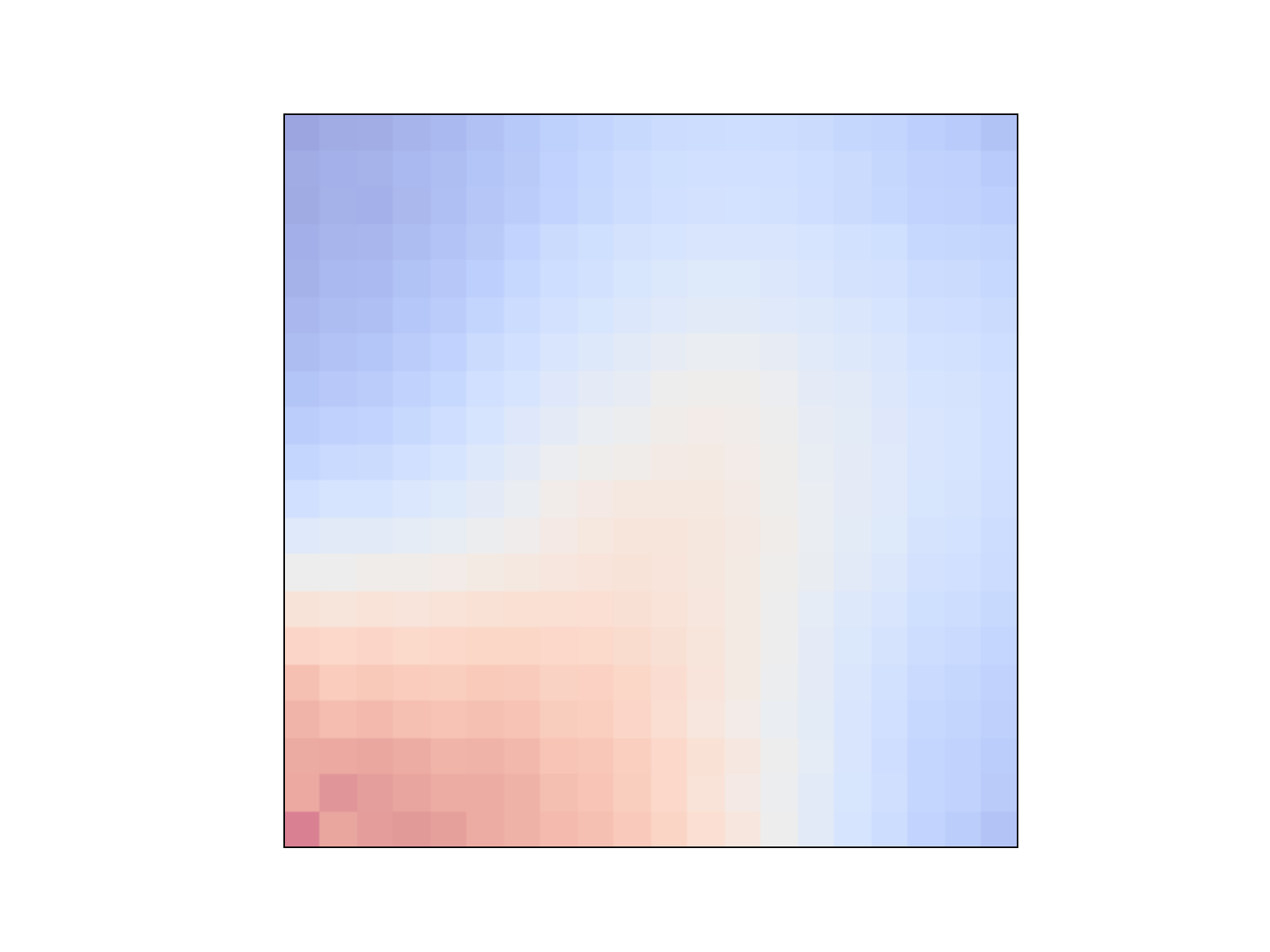}}
\caption{The visualization of the interpolation ($t=21.5$) and the extrapolation ($t=81$ and $t=90$) on the grid network} \label{fig:synth}
\end{figure}

\begin{figure}[h]
\centering
\subfigure[True ($t=21.5$)]{\includegraphics[width=0.32\columnwidth]{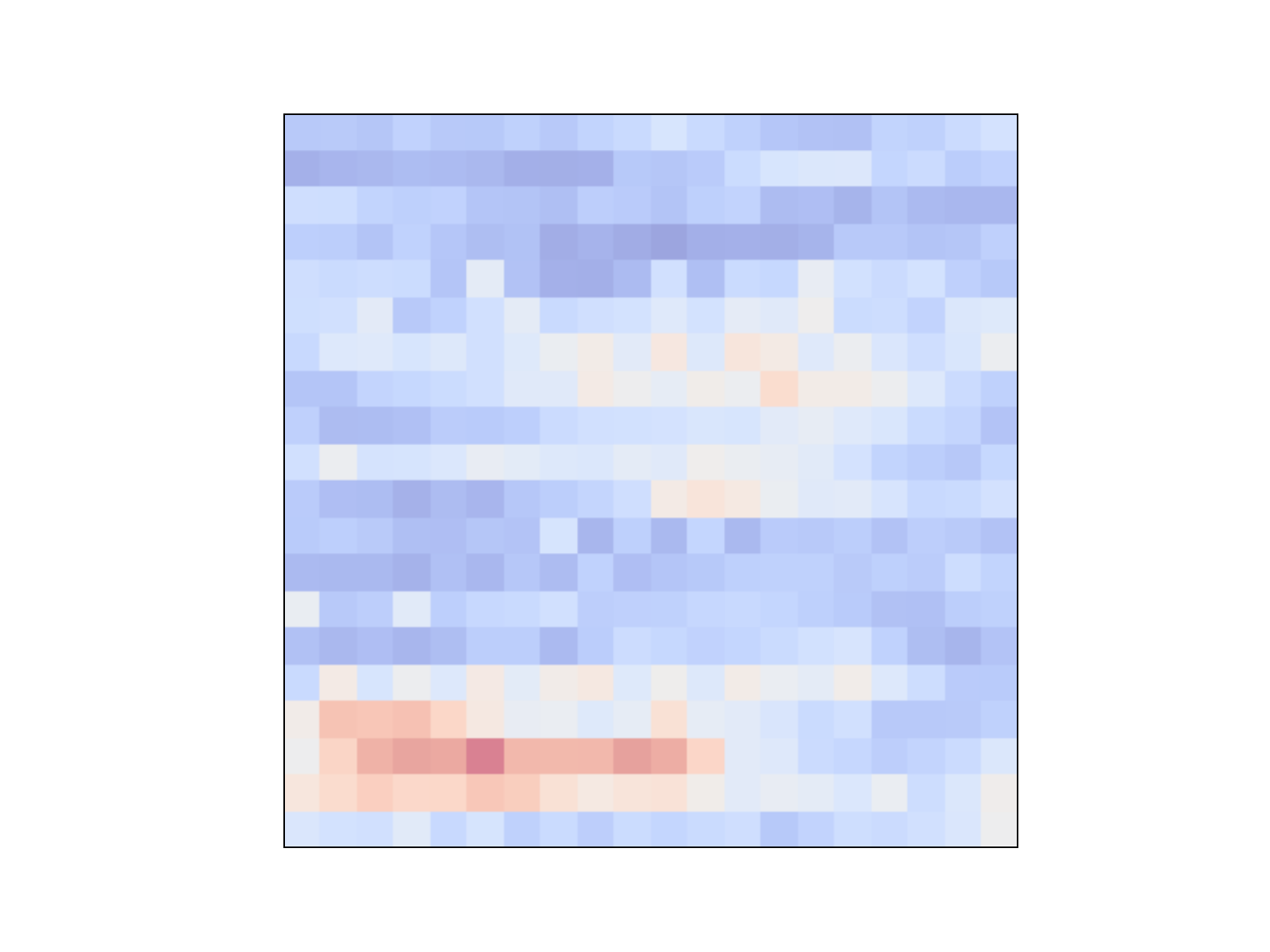}}
\subfigure[True ($t=81$)]{\includegraphics[width=0.32\columnwidth]{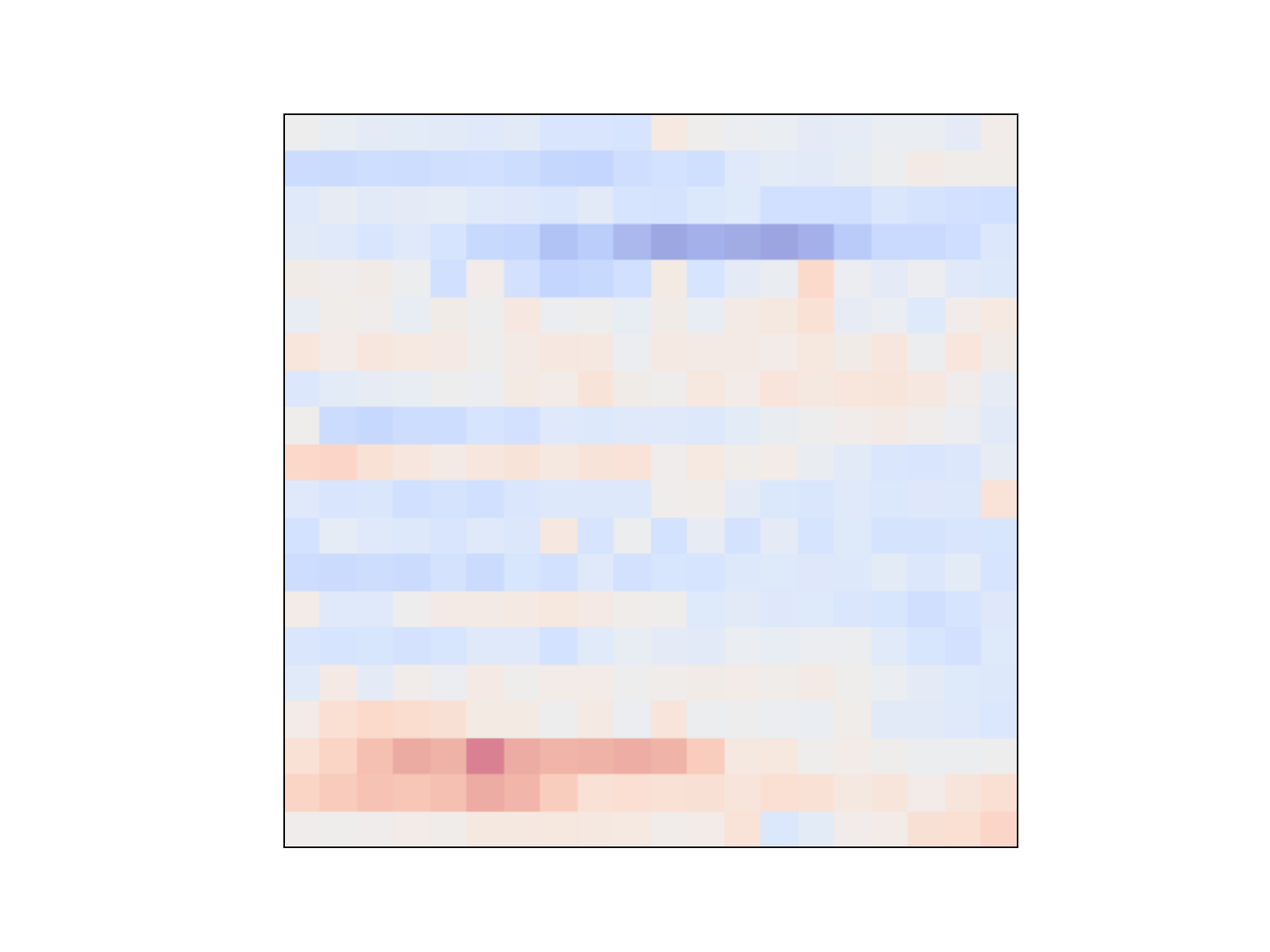}}
\subfigure[True ($t=90$)]{\includegraphics[width=0.32\columnwidth]{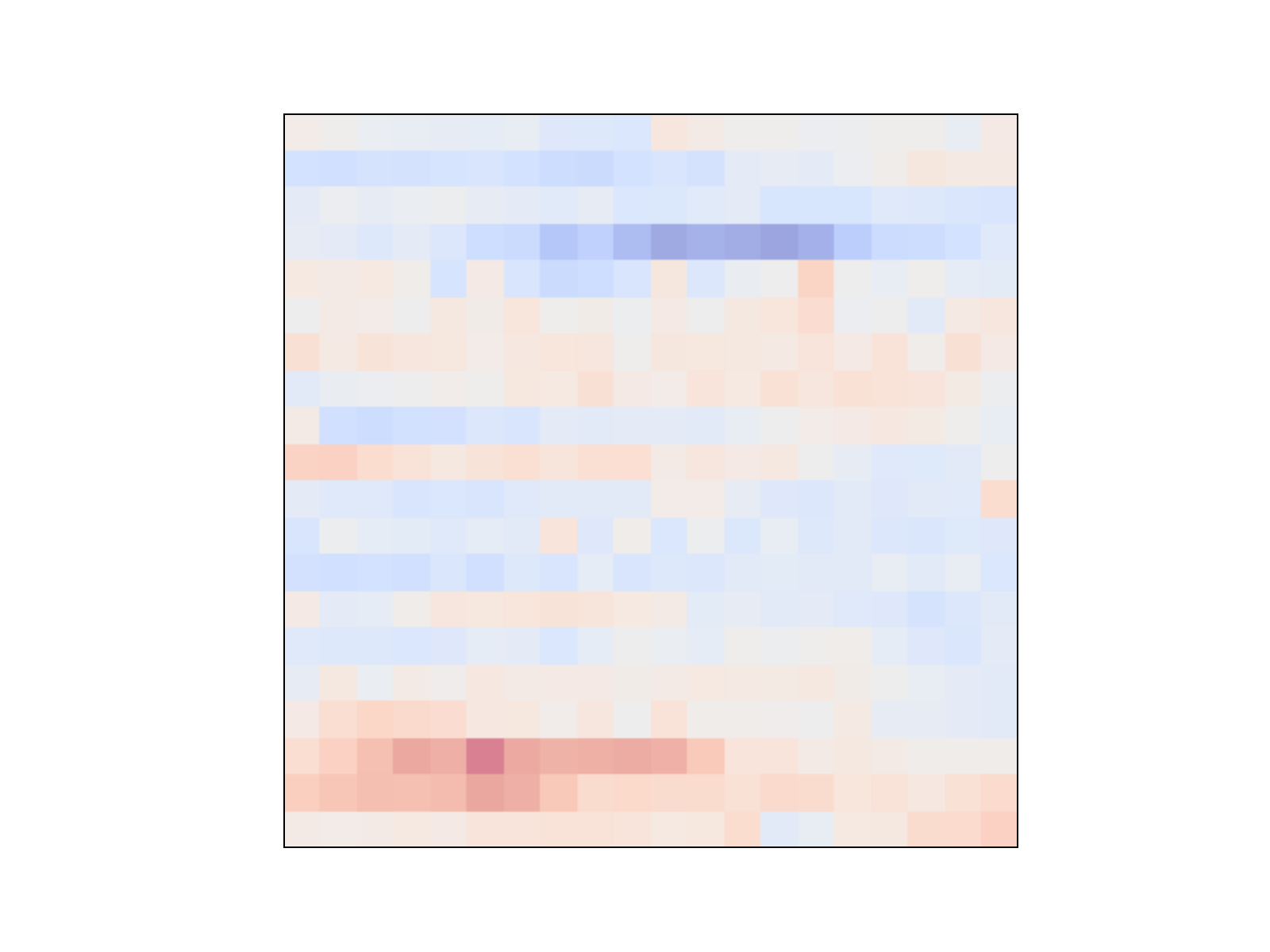}}
\subfigure[NDE ($t=21.5$)]{\includegraphics[width=0.32\columnwidth]{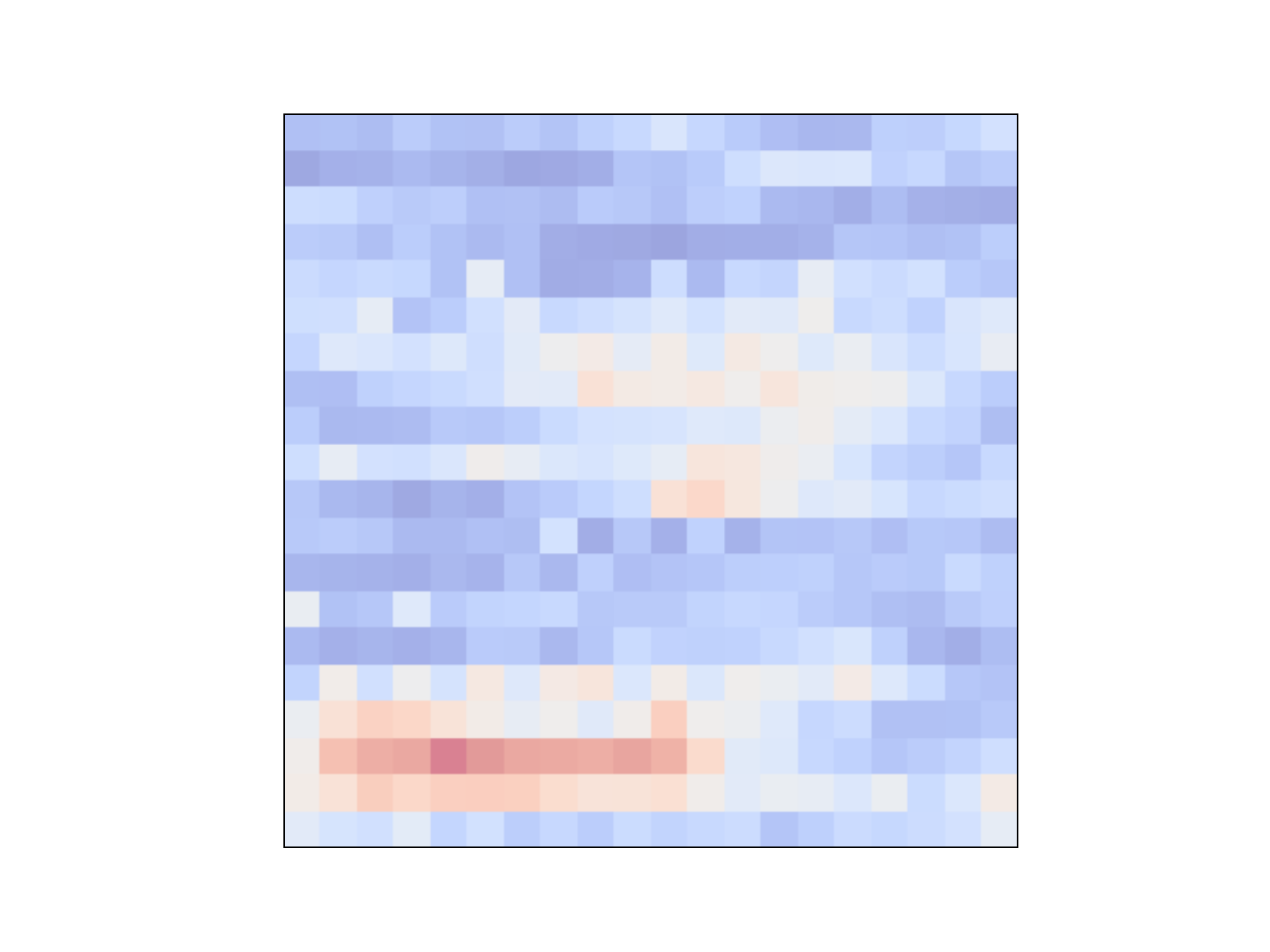}}
\subfigure[NDE ($t=81$)]{\includegraphics[width=0.32\columnwidth]{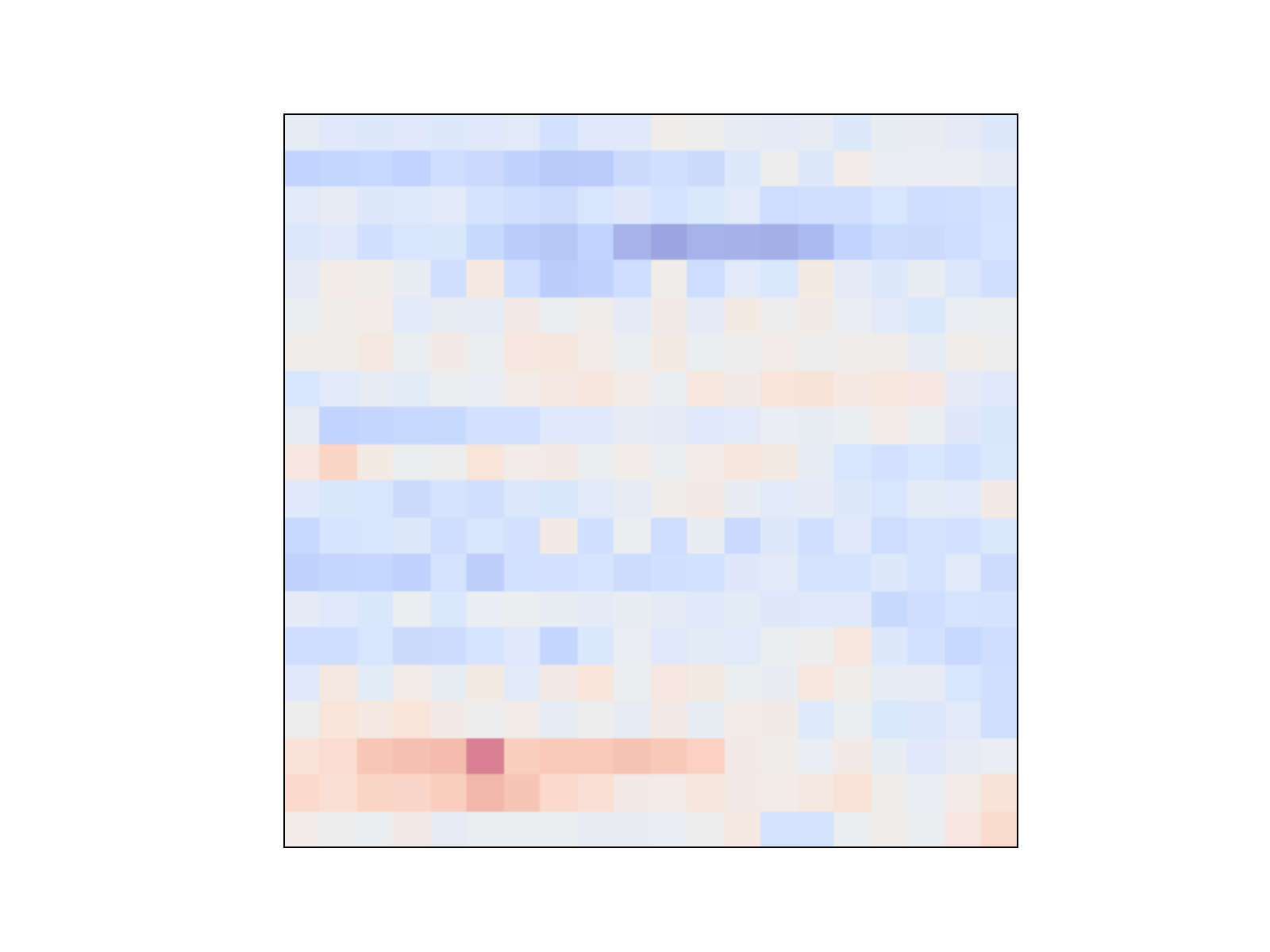}}
\subfigure[NDE ($t=90$)]{\includegraphics[width=0.32\columnwidth]{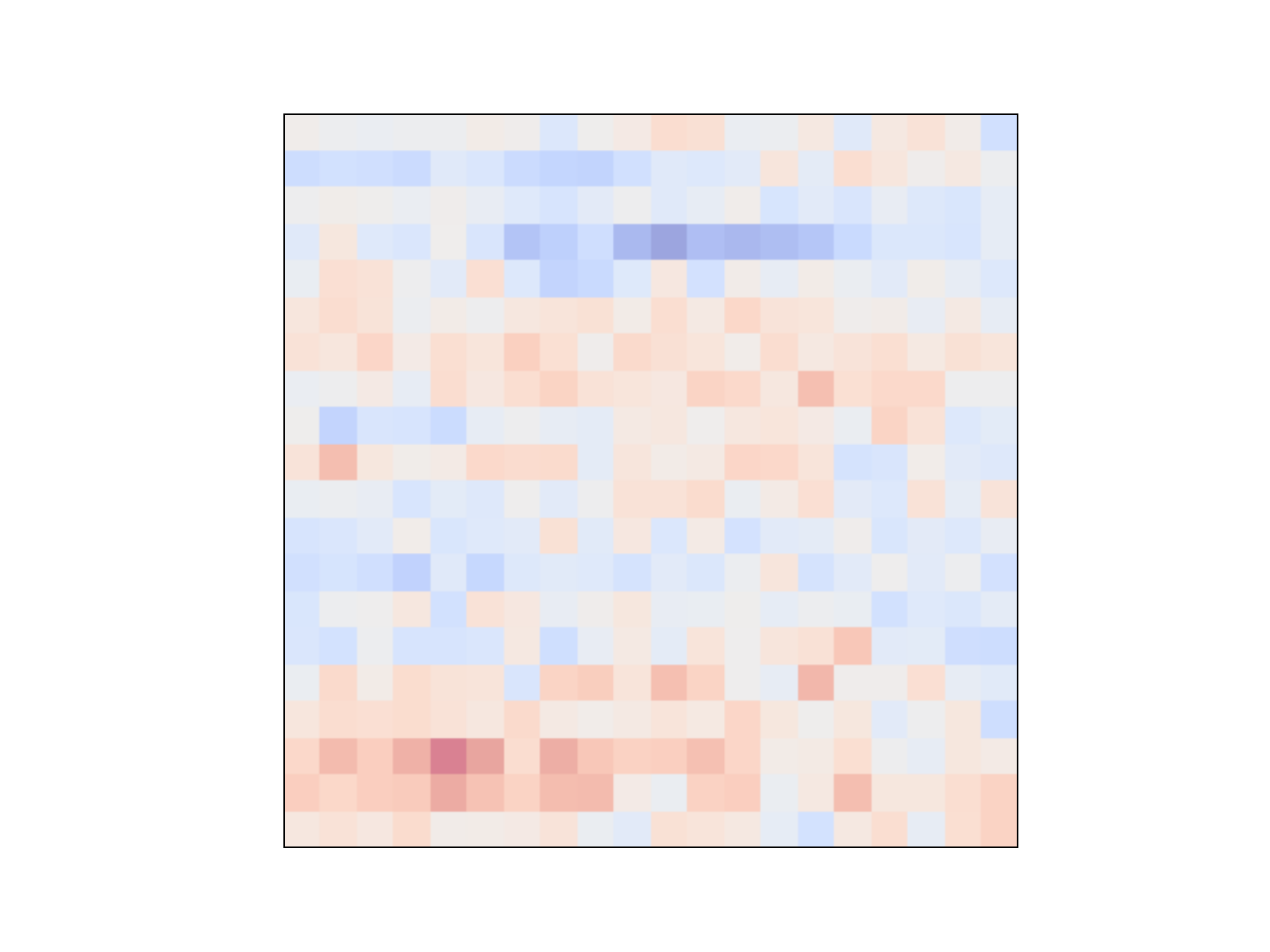}}
\caption{The visualization of the interpolation ($t=21.5$) and the extrapolation ($t=81$ and $t=90$) on the small-world network} \label{fig:synth-sw}
\end{figure}

Fig.~\ref{fig:synth} visualizes some ground-truth and predicted values in the grid network. The true diffusion examples in Fig.~\ref{fig:synth} (a-c) show that there exist three thermal points in (a) but as time goes by, they are diffused over the 2-dimensional space. We note that the time points we use in these figures, i.e., $t \in \{21.5, 81, 90\}$, do not belong to the training set. Therefore, $t=21.5$ means an interpolation and the other two cases mean extrapolations. Our prediction by NDE successfully interpolate and extrapolate the diffusion process as shown in the figures. Fig.~\ref{fig:synth-sw} visualizes for the small-world network and we can observe similar patterns.

\begin{figure*}[!t]
\centering
\includegraphics[width=1\textwidth]{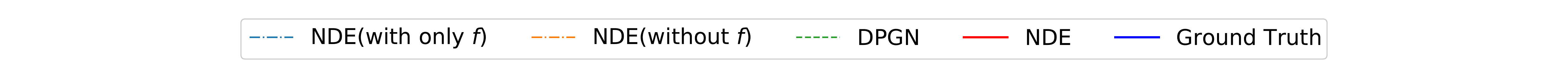}\vspace{-1em}
\subfigure[Node 1 in LA]{\includegraphics[width=0.4\columnwidth]{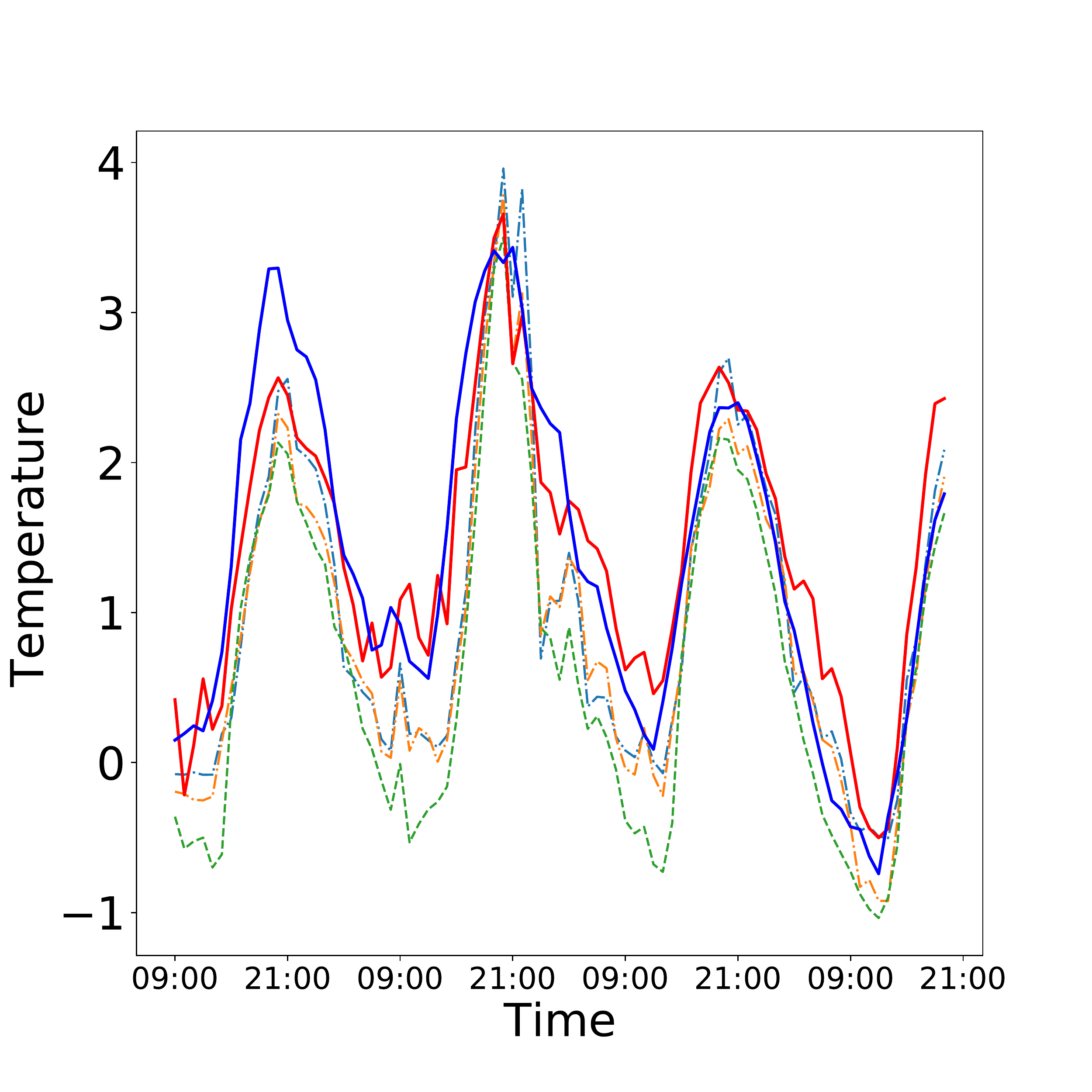}}
\subfigure[Node 2 in LA]{\includegraphics[width=0.4\columnwidth]{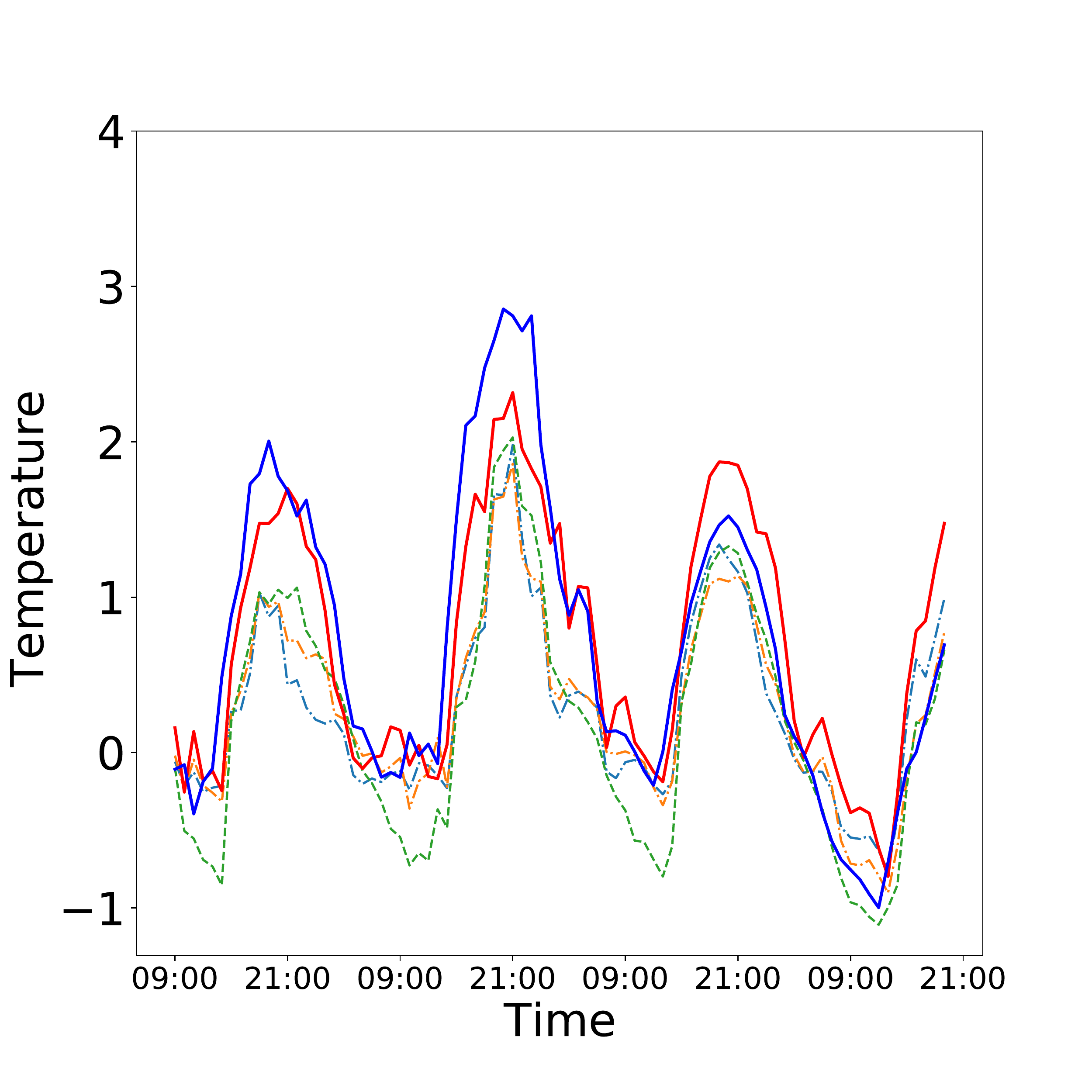}}
\subfigure[Node 1 in SD]{\includegraphics[width=0.4\columnwidth]{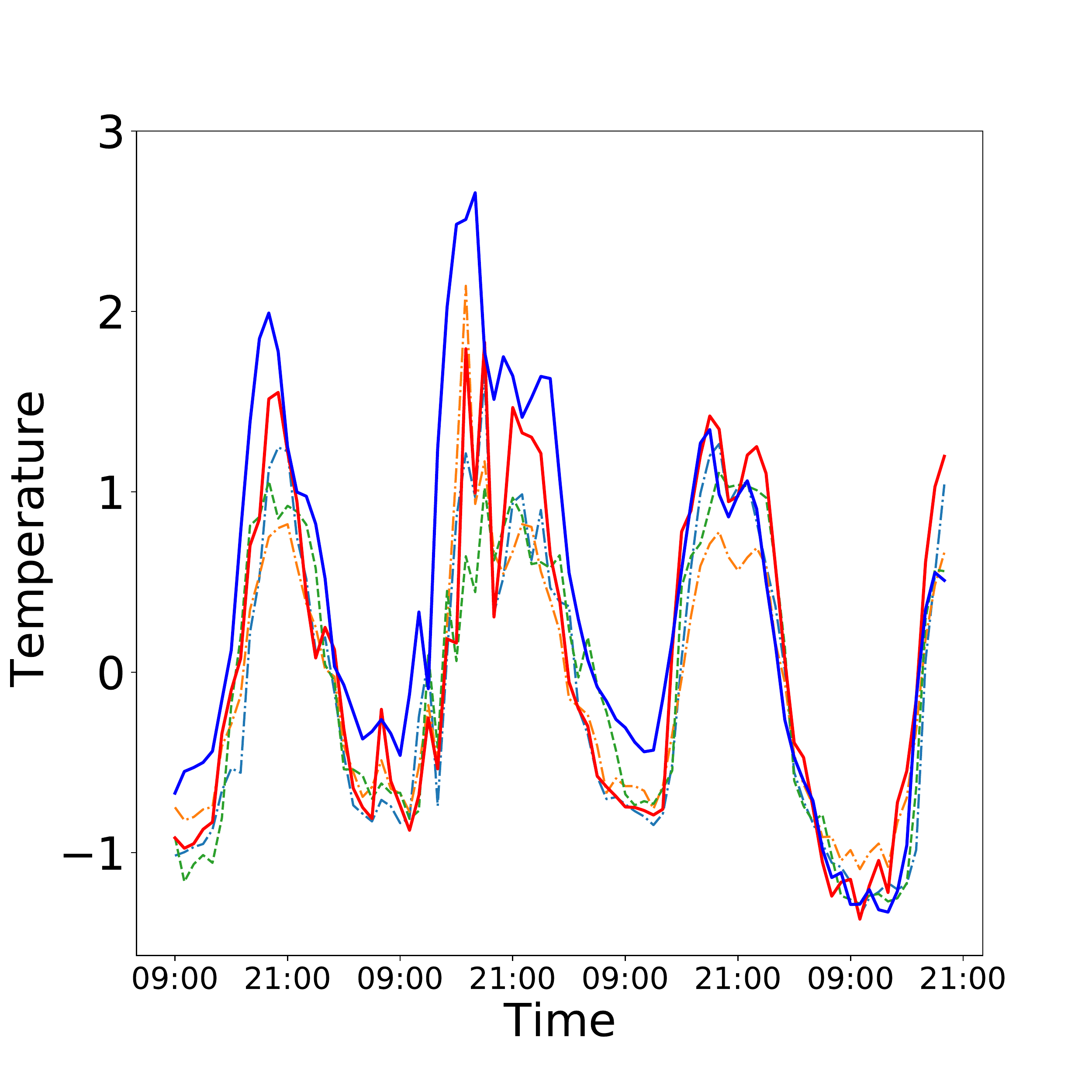}}
\subfigure[Node 2 in SD]{\includegraphics[width=0.4\columnwidth]{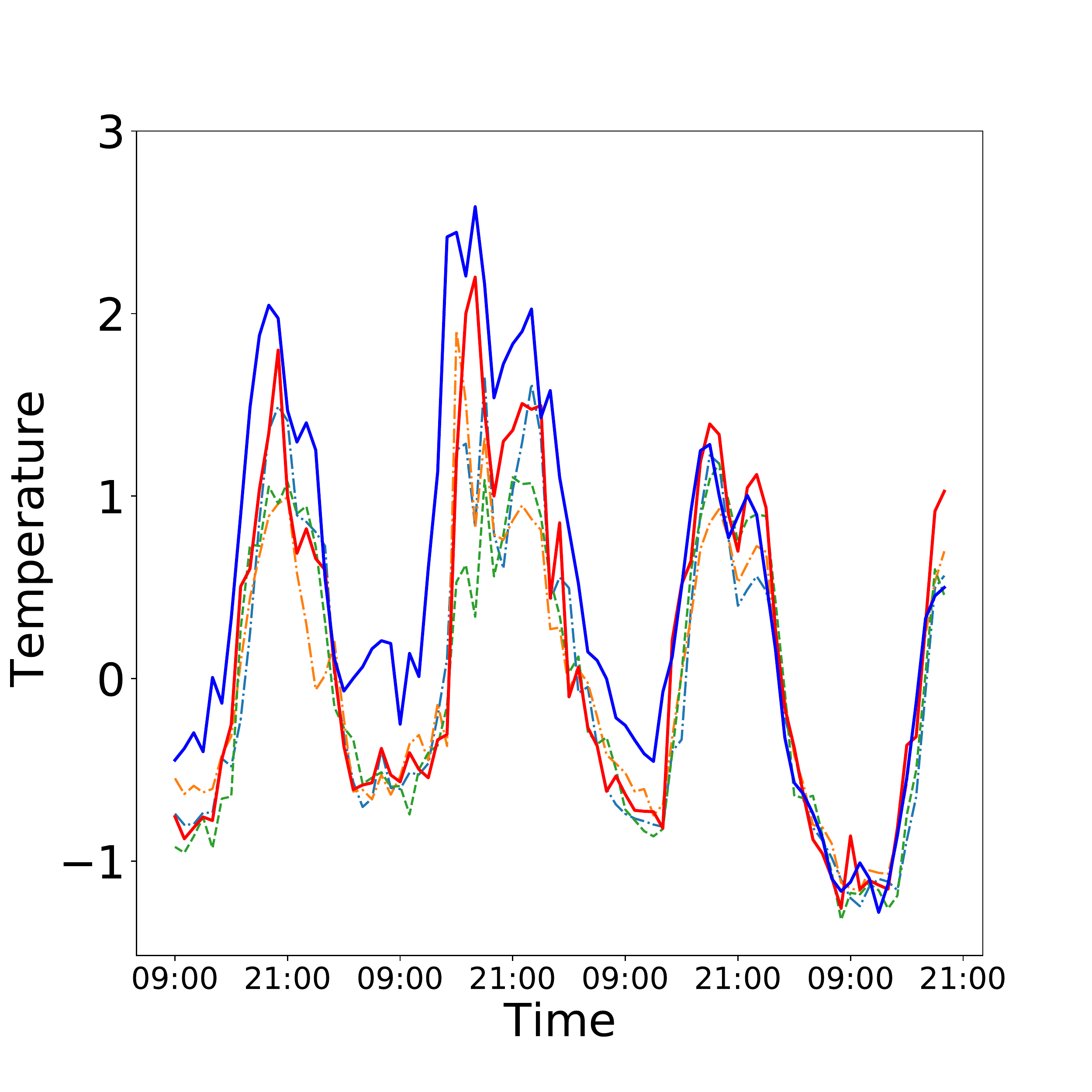}}
\caption{We visualize the ground-truth and predicted values for two nodes of the LA and SD graphs. (a-b) One-step near-surface air temperature forecasting in LA from 2012/07/10 9:00 to 2012/07/14 22:00 (c-d) One-step near-surface air temperature forecasting in SD from 2012/07/10 9:00 to 2012/07/14 22:00}\label{fig:sd_plot}
\end{figure*}

\subsubsection{LA and SD Data} Table~\ref{tbl:lasd} summarizes the forecasting results in the LA and SD datasets. For the one next step forecasting of LA, MLP and RNN-based models show the biggest errors (MSEs) whereas other differential equation-based models show reasonable accuracy. One interesting point is that the combination of RNN and GNN does not significantly improve the accuracy over the only RNN-based models. In comparison with GN-skip, both DPGN and NDCN significantly decrease the errors, i.e., a MSE of 0.5654 for GN-skip vs. 0.4435 of DPGN. However, our NDE further decreases the error to 0.2628, which is about 41\% smaller. For the one next step forecasting in SD, we observe similar patterns with a little smaller margin of 31\% between DPGN and NDE. However, the gap between them is still non-trivial.

For the multi-step forecasting in LA, many baseline methods show unreliable outcomes in comparison with our method, e.g., an MSE of 1.2588 by NDCN vs. 0.7594 by NDE. Except our method, DPGN shows a small error.

For the multi-step forecasting in SD, however, our method does not show the best accuracy. Our NDE is comparable to NDCN and RNN-GNN. The smallest error is achieved by DPGN (although both DPGN and NDE rely on the diffusion equation in their model designs). We think that the training mechanism of PINN used by DPGN is more effective than our training method for SD. Recall that our method is more effective than the PINN mechanism of DPGN for LA, which shows the difficulty of learning diffusion processes to model climate. One more difficulty of SD is that it contains many unpaved regions, as noted in~\cite{seo2019differentiable}.

\begin{table*}[t]
\centering
\setlength{\tabcolsep}{4pt}
\caption{Prediction errors (MSEs) on the LA and SD}\label{tbl:lasd}
\begin{tabular}{ccccccc}
\hline
\multirow{2}{*}{Model} & LA  & SD  & LA  & SD & LA & SD \\ 
 & One-step & One-step & Multi-step & Multi-step & \#Params & \#Params \\   \hline
MLP &  0.6902 $\pm$ 0.0171 & 0.5863 $\pm$ 0.0123  & 1.2766 $\pm$ 0.0143 & 1.0030 $\pm$ 0.0154 & 4,865 & 4,865  \\
RNN &  0.6295 $\pm$ 0.0178 & 0.5411 $\pm$ 0.0231  &  1.2869 $\pm$ 0.0172  &  0.9065 $\pm$ 0.0172 & 29,825 & 29,825\\
LSTM &  0.6723 $\pm$ 0.0074  &  0.5651 $\pm$ 0.0269 &  1.3043 $\pm$ 0.0106  &  0.8811 $\pm$ 0.0225 & 119,105 & 119,105\\
GRU &  0.6486 $\pm$ 0.0049 & 0.5749 $\pm$ 0.0082 &  1.2872 $\pm$ 0.0141 & 0.8852 $\pm$ 0.0084 & 89,345 & 89,345\\
RNN-GNN &  0.6607 $\pm$ 0.0064 &  0.5291 $\pm$ 0.0578 &  1.1844 $\pm$ 0.0727 & 0.7862 $\pm$ 0.0475 & 29,825 & 29,825\\
LSTM-GNN &  0.7007 $\pm$ 0.0008  &  0.5754 $\pm$ 0.0180  & 1.1392 $\pm$ 0.0252 &  0.7954 $\pm$ 0.0110 & 119,105 & 119,105\\
GRU-GNN &  0.6914 $\pm$ 0.0098  &  0.5705 $\pm$ 0.0057 &  1.1722 $\pm$ 0.0726 & 0.7414 $\pm$ 0.0294 & 89,345 & 89,345\\
GN-only & 0.6035 $\pm$ 0.0832 &  0.7007 $\pm$ 0.0848 & 1.3415 $\pm$ 0.1195 &  1.0422 $\pm$ 0.0673 & 45,696 & 45,696\\
GN-skip &  0.5654 $\pm$ 0.1015 & 0.6543 $\pm$ 0.1195 &  1.0257 $\pm$ 0.1912 & 0.9872  $\pm$ 0.2425 & 45,696 & 45,696\\
DPGN &  0.4435 $\pm$ 0.0378 & 0.5149 $\pm$ 0.0831 &  0.8677 $\pm$ 0.1033 & \textbf{0.6714 $\pm$ 0.1106} & 45,696 & 45,696\\
NDCN & 0.5380 $\pm$ 0.0469& 0.5296 $\pm$ 0.0274  & 1.2588 $\pm$ 0.0654 &0.7542 $\pm$ 0.0730 & 75,525 & 75,525 \\\hline
NDE(with only $f$) &  0.3439 $\pm$ 0.0922  &  0.6009 $\pm$ 0.2001 & 0.8635 $\pm$ 0.0267 & 0.7960 $\pm$ 0.0854 & 4,851 & 9,889\\
NDE(without $f$) & 0.3493 $\pm$ 0.0488 &  0.5522 $\pm$ 0.0582 & 0.9730 $\pm$ 0.1266 & 0.8731 $\pm$ 0.0236 & 492 & 1,454\\  
\textbf{NDE} & \textbf{0.2621 $\pm$ 0.0026} & \textbf{0.3561 $\pm$ 0.0055} & \textbf{0.7594 $\pm$ 0.0225} & 0.7301 $\pm$ 0.0048 & 4,894 & 9,934\\
\hline
\end{tabular}
\end{table*}


Fig.~\ref{fig:sd_plot} shows the ground-truth and predicted values by time. In these figures, we compare only the highly performing differential equation-based methods. With the human visual evaluation in the figures, our NDE shows the best matches with the ground-truth values. DPGN's predictions are clearly worse than those of our method.

\subsubsection{NOAA Data} For this dataset which does not have edge features, DPGN, GN-only and GN-skip cannot be tested because they require edge features. In NOAA, differential equation-based methods show a good match. MLP and RNN-based methods show much larger errors than those of NDCN and NDE. Among RNN-based models, LSTM shows relatively smaller errors. However, NDCN shows much smaller errors than them, and our NDE shows the smallest errors among all methods.

\begin{table}[ht]
\centering
\setlength{\tabcolsep}{4pt}
\caption{Prediction errors (MAEs) on NOAA}\label{tbl:noaa}
\begin{tabular}{cccc}
\hline
Model & One-step & Multi-step & \#Params\\ \hline
MLP & 0.4629 $\pm$ 0.0187 & 3.2582 $\pm$ 0.5518 & 4,673 \\
RNN &  0.6783 $\pm$ 0.0675 & 4.5232 $\pm$ 0.2807 & 29,313 \\
LSTM & 0.4822 $\pm$ 0.0356 & 2.4247 $\pm$ 0.0906 & 117,057 \\
GRU &  0.5930 $\pm $ 0.0838 & 2.1406 $\pm$ 0.0409 & 87,809  \\
RNN-GNN &  0.7169 $\pm$ 0.1131 &  6.5125 $\pm$ 1.4519 & 29,313 \\
LSTM-GNN & 0.4631 $\pm$ 0.0292 &  2.3955 $\pm$ 0.7086 & 117,057 \\
GRU-GNN &  0.5020 $\pm$ 0.1098 &  1.9906 $\pm$ 0.1261 & 87,809 \\
NDCN &  0.3151 $\pm$ 0.0122 & 2.2967 $\pm$ 0.0415 & 36,657 \\\hline
NDE(with only $f$) & 0.3340 $\pm$ 0.0265  & 1.8039 $\pm$ 0.1017 & 85,089  \\
NDE(withouf $f$) &  0.3245 $\pm$ 0.0161 & 1.8890 $\pm$ 0.0565 & 37,713 \\
\textbf{NDE} & \textbf{0.2975 $\pm$ 0.0062} & \textbf{1.6337 $\pm$ 0.0467} & 53,585 \\
\hline
\end{tabular}
\end{table}

\subsection{Ablation and Sensitivity Studies}
With the LA and SD datasets, we conduct ablation studies. First, we compare the heat capacity generation methods in Table~\ref{tbl:ablation1}. In all cases, as shown, the heat capacity generation by edge class shows the smallest errors, followed by the single coefficient and the fixed coefficient methods.

\begin{table}[ht]
\centering
\setlength{\tabcolsep}{4pt}
\caption{Sensitivity w.r.t. the heat capacity generation method for one-step prediction}\label{tbl:ablation1}
\begin{tabular}{ccc}
\hline
Heat Capacity Generation Method & LA & SD \\ \hline
FC &  0.3657 $\pm$ 0.0117 & 0.7701 $\pm$ 0.1164   \\
SC &  0.2901 $\pm$ 0.0096 & 0.5863 $\pm$ 0.0550 \\
HM & 0.5063 $\pm$ 0.1031 &  0.6033 $\pm$ 0.0382 \\
EC & \textbf{0.2621 $\pm$ 0.0026} &  \textbf{0.3561 $\pm$ 0.0055}  \\
\hline
\end{tabular}
\end{table}

We also carried out sensitivity studies by varying $P$, i.e., varying how much past information we feed into the model. In Table~\ref{tbl:p_history}, it shows that climate modeling does not require long history. The diffusion equation also requires only $\bm{H}(t)$ to derive $\frac{d \bm{H}(t)}{d t}$, which means past history is not needed in the diffusion equation. Our experimental results also show similar patterns in LA and SD.

\begin{table}[ht]
\centering
\setlength{\tabcolsep}{4pt}
\caption{Sensitivity w.r.t. $P$}\label{tbl:p_history}
\begin{tabular}{ccc}
\hline
$P$ & LA & SD \\ \hline
1 &  0.2621 $\pm$ 0.0026 & 0.3561 $\pm$ 0.0055\\
3 & 0.3332 $\pm$ 0.0459 & 0.4495 $\pm$ 0.0899\\
5 & 0.3229 $\pm$ 0.0868 & 0.4289 $\pm$ 0.0576 \\
\hline
\end{tabular}
\end{table}

Table~\ref{tbl:hidden} shows the performance by varying the hidden dimension size $D$. We test up to $D=32$ and $D=32$ shows the smallest errors. In particular, the performance gap between $D=16$ and $D=32$ is large in SD and NOAA.

\begin{table}[ht]
\centering
\setlength{\tabcolsep}{4pt}
\caption{Sensitivity w.r.t. the hidden dimension size $D$}\label{tbl:hidden}
\begin{tabular}{cccc}
\hline
Size of $D$ & LA & SD & NOAA\\ \hline
    8 & 0.2787 $\pm$ 0.0049 & 0.5223 $\pm$ 0.0492 & 0.4355 $\pm$ 0.0092\\
16 & 0.2628 $\pm$ 0.0132 & 0.5491 $\pm$ 0.0671 & 0.3789 $\pm$ 0.0135\\
32 & 0.2621 $\pm$ 0.0026 & 0.3561 $\pm$ 0.0055 & 0.2975 $\pm$ 0.0062\\
\hline
\end{tabular}
\end{table}

In Fig.~\ref{fig:sd_plot} and other main result tables, the two ablation study models, i.e., NDE(without $f$) and NDE(with only $f$), do not produce as good predictions as those of the full model, NDE. One more point worth mentioning is that NDE(with only $f$) sometimes requires a large model for $f$ since it relies only on $f$, e.g., 85,089 parameters in NOAA by NDE(with only $f$) vs. 53,585 by NDE.

\subsection{Model Efficiency Analyses}
\begin{figure}[h!]
\centering
\subfigure[One-step prediction]{\includegraphics[width=0.49\columnwidth]{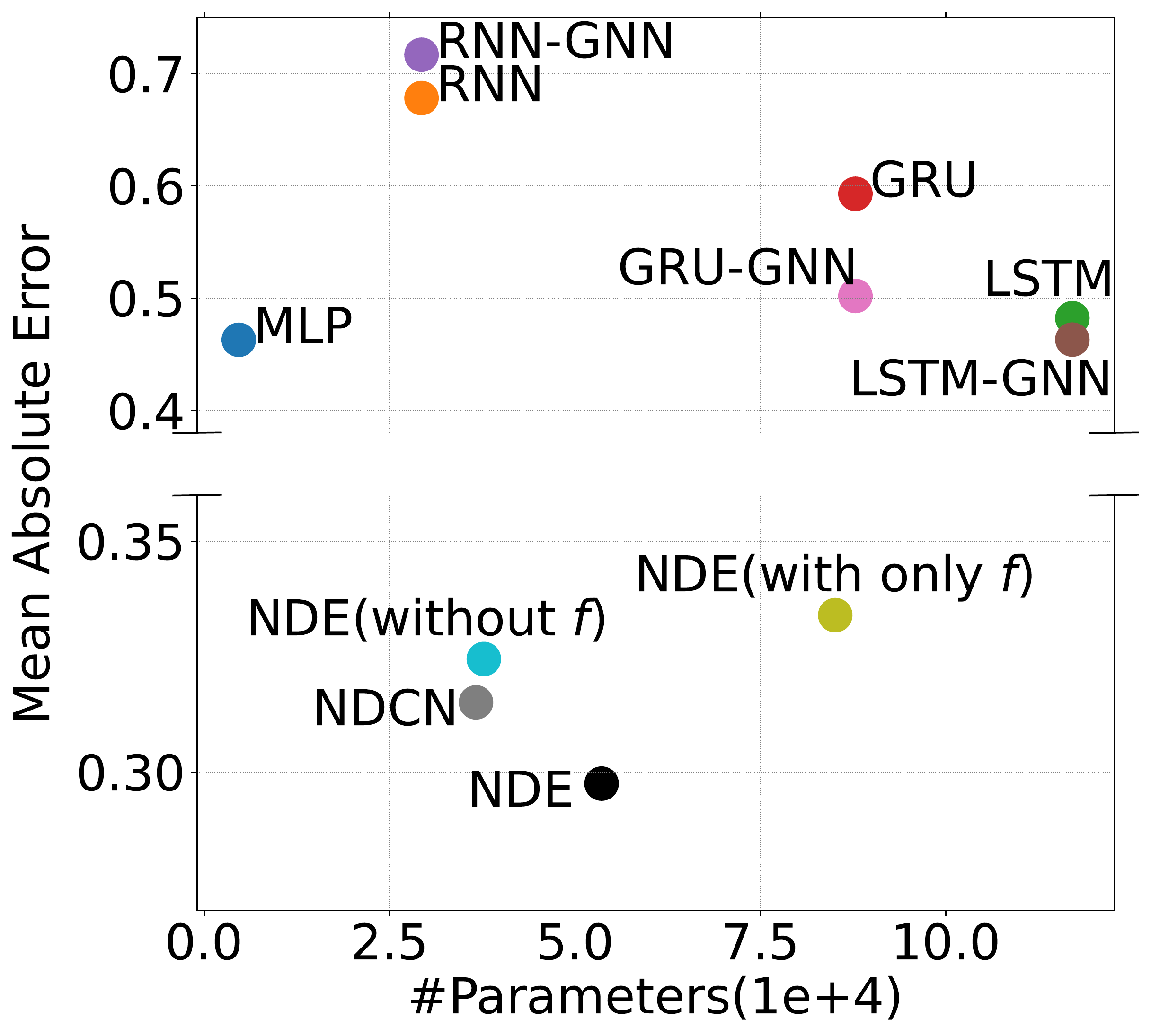}}
\subfigure[Multi-step prediction]{\includegraphics[width=0.49
\columnwidth]{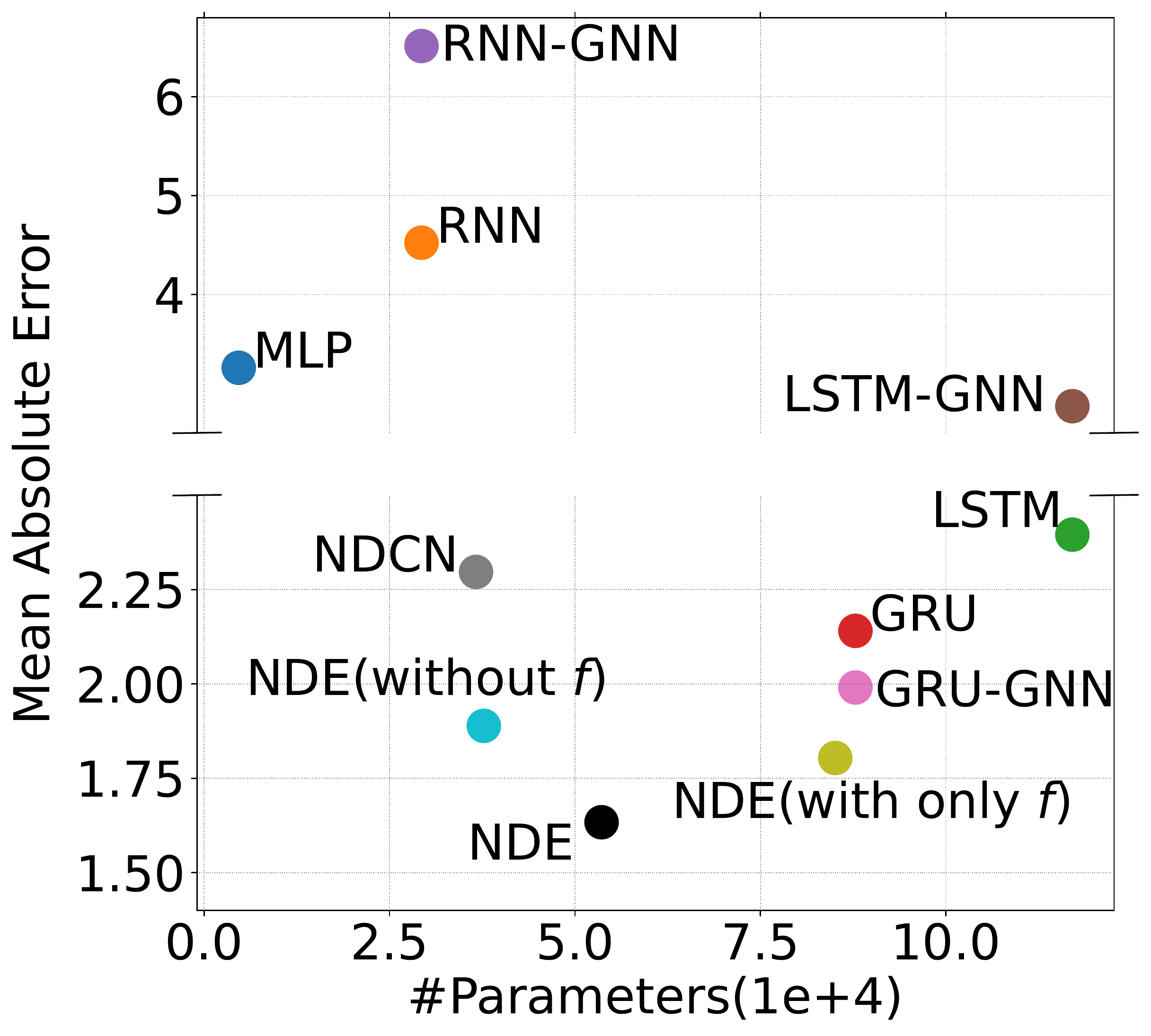}}
\caption{Model efficiency in NOAA. The bottom left corner is preferred.} \label{fig:parameter}
\end{figure}

Fig.~\ref{fig:parameter} shows the number of parameters and the error. Models at the bottom left corner in this figure are preferred. Our NDE and its variants are located around the bottom left corner.

In general, RNN-based models show low efficiencies in the figure, followed by RNN-GNN, GRU-RNN, and NDCN. One more interesting point is that our ablation study models also show good efficiencies in general.

\section{Conclusion}
In this paper, we tackled the problem of one-step and multi-step climate factor forecasting. The diffusion equation with uncertainty modeling is used in our research. To this end, our presented method learns i) the heat capacity generation method and ii) the neural network-based uncertainty model. In the end, these two modules are combined with the diffusion equation. Our comprehensive experiments with synthetic and real-world datasets show the best efficiency of our method, NDE. In general, NDE achieves the best (or nest-best) accuracy with a relatively smaller number of parameters and GPU memory space complexity during inference.

In the future, we will study a more principled method to model uncertainty. One possible approach is to use stochastic differential equations (SDEs)~\cite{NEURIPS2019_770f8e44,pmlr-v119-kong20b}.

\section*{Acknowledgement}
Noseong Park (noseong@yonsei.ac.kr) is the corresponding author. This work was supported by the Yonsei University Research Fund of 2021, and the Institute of Information \& Communications Technology Planning \& Evaluation (IITP) grant funded by the Korean government (MSIT) (No. 2020-0-01361, Artificial Intelligence Graduate School Program (Yonsei University), and No. 2021-0-00155, Context and Activity Analysis-based Solution for Safe Childcare).

\bibliographystyle{IEEEtran}
\bibliography{ref}

\end{document}